\documentclass[default,iicol]{sn-jnl}

\jyear{2022}%

\usepackage{amsmath}
\usepackage{xspace}
\usepackage{flushend}
\usepackage{color}
\usepackage{natbib}
\usepackage[capitalise]{cleveref}
\usepackage{ulem}
\usepackage{epsfig}
\usepackage{epstopdf}
\usepackage[justification=centering]{caption}
\usepackage{float}

\crefname{algorithm}{Alg.}{Algs.}
\crefname{figure}{Figure}{Figures}
\crefname{table}{Table}{Tables}
\crefname{equation}{Equation}{Equation}

\theoremstyle{thmstyleone}%
\theoremstyle{thmstyletwo}%

\theoremstyle{thmstylethree}%

\newcommand{\PosReals}{\mathbb{R}_+}
\newcommand{\NNegReals}{\mathbb{R}_{\geq 0}}
\newcommand{\PosNaturals}{\mathbb{N}_+}
\newcommand{\Naturals}{\mathbb{N}}

\newcommand{\activityspace}{\mathcal{A}}

\newcommand{\eventset}{\mathcal{E}}
\newcommand{\entityset}{\mathcal{C}}
\newcommand{\eventfeatureset}{\mathcal{F}_{event}}
\newcommand{\entityfeaturespace}{\mathcal{F}_{entity}}
\newcommand{\freq}{\vec{f}}

\newcommand{\fw}{CAPiES\xspace}

\newcommand{\size}[1]{\left\vert #1 \right\vert}
\newcommand{\smallsize}[1]{\vert #1 \vert}

\newcommand*{\ceil}[1]{\left\lceil #1 \right\rceil}

\newcommand{\additem}[1]{\textcolor{olive}{#1}}
\newcommand{\edititem}[2]{\textcolor{blue}{#1} \textcolor{blue}{\sout{#2}}}
\newcommand{\remitem}[1]{\textcolor{red}{\sout{#1}}}

\renewcommand{\additem}[1]{#1}
\renewcommand{\edititem}[2]{#1}
\renewcommand{\remitem}[1]{}

\begin{document}

\title[Clustering-based Aggregations for Prediction in Event Streams\remitem{ of Shopper Behaviour}]{Clustering-based Aggregations for Prediction in Event Streams\remitem{ of Shopper Behaviour}}  

\author*[1]{\fnm{Yorick} \sur{Spenrath}}\email{y.spenrath@tue.nl}

\author[1]{\fnm{Marwan} \sur{Hassani}}\email{m.hassani@tue.nl}

\author[1]{\fnm{Boudewijn F.} \sur{Van Dongen}}\email{b.f.v.dongen@tue.nl}

\affil[1]{\orgdiv{Process Analytics, Department of Mathematics and Computer Science}, \orgname{Eindhoven University of Technology}, \orgaddress{\city{Eindhoven}, \country{The Netherlands}}}


\abstract{Predicting the behaviour of shoppers provides valuable information for retailers, such as the expected spend of a shopper or the total turnover of a supermarket. The ability to make predictions on an individual level is useful, as it allows supermarkets to accurately perform targeted marketing. However, given the expected number of shoppers and their diverse behaviours, making accurate predictions on an individual level is difficult. This problem does not only arise in shopper behaviour, but also in various business processes, such as predicting when an invoice will be paid. In this paper we present \fw, a framework that focuses on this trade-off in an online setting. By making predictions on a larger number of entities at a time, we improve the predictive accuracy but at the potential cost of usefulness since we can say less about the individual entities. \fw is developed in an online setting, where we continuously update the prediction model and make new predictions over time. We show the existence of the trade-off in an experimental evaluation in two real-world scenarios: a supermarket with over 160 000 shoppers and a paint factory with over 171 000 invoices.}

\keywords{Predictive Process Mining, Stream Analysis, Clustering, Shopper Behaviour}



\maketitle

\section{Introduction}\label{sec:intro}
\begin{figure*}
    \centering
    \includegraphics[width=\textwidth]{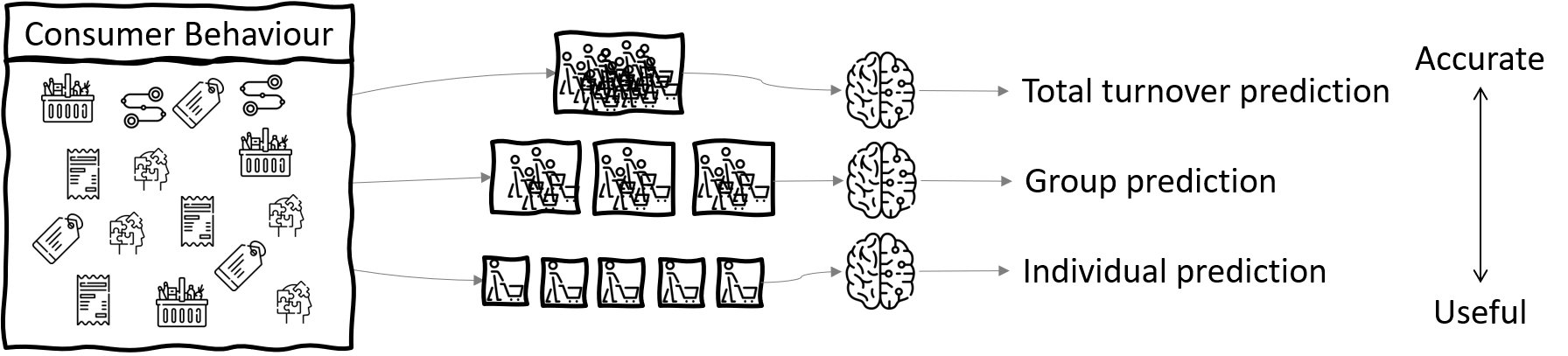}
    \caption{Overview of the problem. Making predictions for individual consumers is more useful, but less accurate. Making predictions for all consumers (i.e. the entire retailer) is easier, but not useful on individual consumers. This paper balances the two by making predictions using groups of consumers.}
    \label{fig:problem}
\end{figure*}

Knowing the future behaviour of shoppers is important to help retailers plan ahead \cite{Chamberlain2017}. One way to do so is by predicting how shoppers will behave on an individual level. Knowing which shoppers are expected to increase or decrease their spending allows retailers to apply more targeted marketing strategies. Unfortunately, the human nature of shoppers makes it difficult to accurately predict on an individual level. Two shoppers can behave similar for some time, but then quite different in the next week. This problem becomes extra difficult using product level data, which is often sparse \cite{Ishigaki2018}. Another way to make predictions is by taking all shoppers together. This can then be used to predict the total store turnover based on the turnover of the past weeks. This improves the accuracy of the predictions as we effectively remove outliers from individual consumers, but it also reduces the information we get about the individuals. Data analyzed at this high level of aggregation can also be used to find colloquial regions that explain purchase differences across different geographical locations \cite{Mishra2017}. This trade-off is schematically presented in Figure \ref{fig:problem}.

Similar considerations exist in different settings, such as the ordering process of a paint factory, where we want to predict the time between receiving and paying an invoice. Other work regarding prediction of human-based behaviour is discussed in \cite{DeCnudde2020}.

In this paper we aim to strike a balance between accuracy and usefulness. Instead of making predictions on \textit{individual} entities we make predictions on \textit{groups} of entities. The advantage is an increase in accuracy with respect to making predictions for individual entities as we remove the effect of outliers on the prediction. At the same time, we increase the usefulness of the prediction with respect to the prediction on all entities together. This is because the predictions are on a limited number of entities at a time. \additem{We call this framework `Clustering-based Aggregations for Predictions in Event Streams', or \fw for short.}

\begin{figure*}
    \centering
    \includegraphics[width=\textwidth]{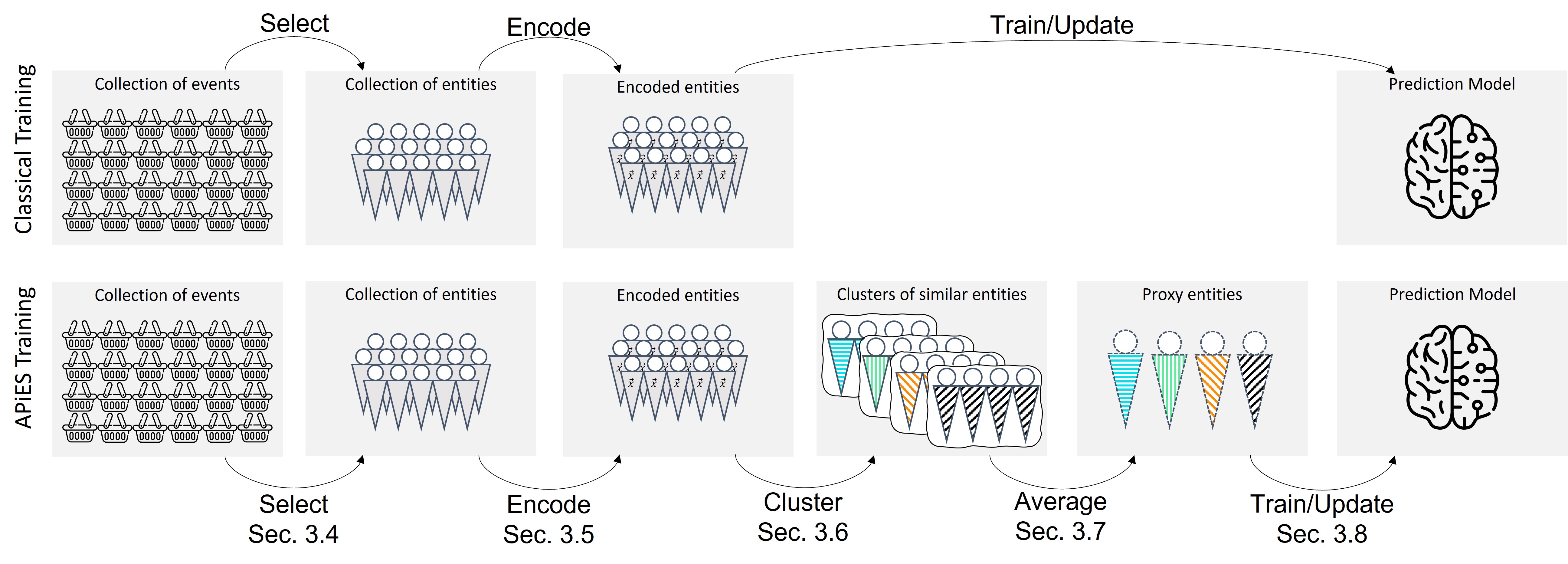}
    \caption{Overview of \fw in the training step}
    \label{fig:overview:training}
\end{figure*}

\begin{figure*}
    \centering
    \includegraphics[width=.833\textwidth]{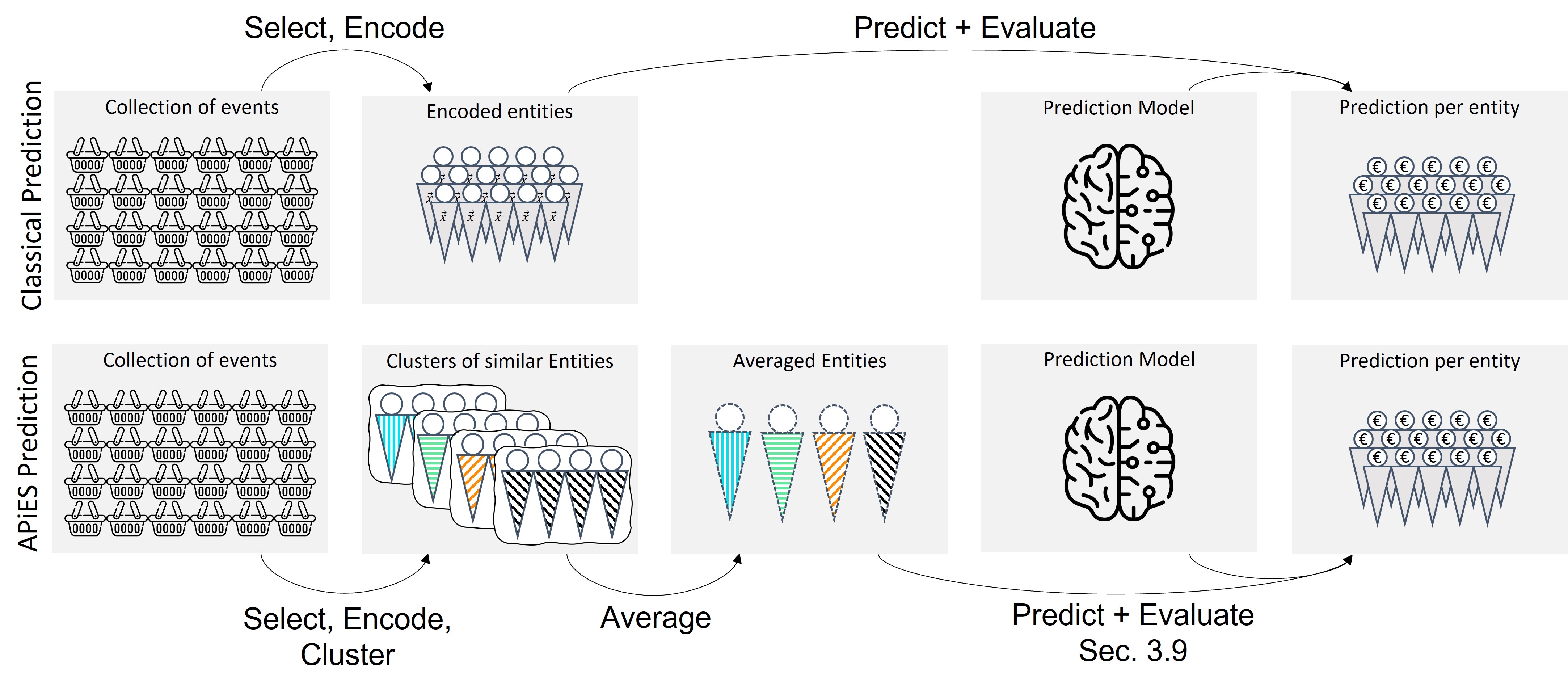}
    \caption{Overview of the framework in the prediction step}
    \label{fig:overview:prediction}
\end{figure*}

\additem{The outline of our method is shown in \cref{fig:overview:training,fig:overview:prediction}, showing the training and prediction phase respectively. Given a set of events, we want to make predictions about entities that relate to these events. In the training phase, our \fw extends the traditional setup of selecting and encoding entities and training the prediction model. In \fw we add two intermediate steps: clustering and averaging. Instead of using the encoded entities to directly update the model, we first partition the set of entities into smaller groups. The idea is that we combine entities that are similar, in other words we apply clustering based on the encoded entities. Next, we compute a \textit{proxy} entity that represents each cluster. The encoding of this proxy entity is obtained by averaging the encoding of the individual entities in the respective cluster. We then use these proxy entities in the training step to learn the prediction model. These two steps are also followed in the prediction phase. After selecting, potentially different, entities and encoding them, the entities are clustered and proxy entities are created which are used in the prediction instead of the individual entities. We then assign the predictions of the proxy entities to each individual entity in the respective cluster.}

\additem{In our framework, a single parameter controls how many entities are added to a cluster in the clustering step of the training and prediction phase. This introduces a trade-off. On the one hand, fewer entities per cluster means that each proxy entity is closer to the individual entity-level and we have more useful predictions. On the other hand, more entities per cluster improves the prediction accuracy of the proxy entity, as the proxy entity is better averaged.}

\additem{Note that the framework does not depend on the specification of the clustering (algorithm, distance metric, other parameters) and the model or its training (model type, classification/regression, train/validation splits, hyper-parameter optimization). Furthermore, while the above description assumes a single training of the model, the same idea can be applied to multiple training phases of a model, for example in a streaming setting where the model updates and entity predictions occur at regular intervals.}

\edititem{In this paper, we additionally apply \fw in a streaming setting, where we discover clusters of entities, update the prediction model, and make predictions at regular intervals. This adaption to a streaming setting is a design choice with generalizability in mind, the use cases presented in this paper are also analyzed in a streaming setting. Note however that \fw is not limited to this, the idea of clustering entities before training/prediction applies to static settings as well.}{
We apply our framework in a streaming setting, at regular intervals we discover clusters of entities to update the prediction model.} We as such make the following contributions: 1) we propose a framework to overcome the loss in prediction accuracy for diverse entities by making the predictions on carefully selected clusters of entities, 2) we do so in a streaming environment, and 3) we show its effectiveness in two different real-world datasets, one from a supermarket (where entities are shoppers) and one from a paint factory ordering process (where entities are invoices).

The content in this paper advances our the findings of our earlier work\footnote{An anonymized version of this earlier work can be found at \url{https://anonymous.4open.science/r/ITEM-CAPiES/Original\%20Version.pdf}}. With respect to that paper we make the following extensions: 1) instead of a framework tailored towards supermarket shoppers specifically, we propose a generic framework that works on a variety of datasets. This generic framework, \fw, is expressed in a more formal manner. 2) Furthermore we change the parameter $k$ (indicating the number of clusters searched for) to $\rho$ (the average number of entities in a cluster), which generalizes better. 3) we apply \fw to the dataset used in the original work, as well as to a completely new dataset. This new dataset comes from a different domain, describing the ordering process in a paint factory. In the current paper, \cref{sec:framework} is fundamentally different from the corresponding section in the original version. Next to this, \cref{sec:results:shoppers} contains updated results and discussions of the existing dataset, \cref{sec:results:paintfactory} describes the results of experiments on the new dataset and \cref{sec:results:conclusion} provides an extended discussion on the results of both experiments together. All other sections have been updated to reflect the new framework and additional experiments.

The rest of this paper is organized as follows: \edititem{we first compare the above overview to existing literature in \cref{sec:related}. Next, in \cref{sec:framework} we explain the details of \fw, each step consisting of a generic formalization, how it applies to a working example of supermarket data, and how that specific step relates to existing literature. Following this, we extensively evaluate \fw in experimental settings using two real world datasets in \cref{sec:experiments}. We conclude the paper in \cref{sec:conclusion}}{we first present our framework in Section \ref{sec:framework} and evaluate it in Section \ref{sec:experiments}. We then discuss how this paper relates to existing work in Section \ref{sec:related}. Finally, we conclude the paper in Section \ref{sec:conclusion}.}

\section{\edititem{Related Work}{}}\label{sec:related}
In this section we discuss related work, but we limit this to comparisons with \fw in general and making predictions in event based use cases. A detailed comparison between specific steps in \fw and existing literature is incorporated in \cref{sec:framework}.

One class of supervised learning is called `bucketing'. In bucketing, datapoints from a training set are first clustered using some clustering method, and a separate machine learning model is trained on each cluster. This approach is also extensively used in predictive process mining. Examples of such works are \cite{Francescomarino2019} (offline) and \cite{difrancescomarino2018} (online). Our approach is different in the sense that we do not train one model per cluster, but train and update a single model using datapoints that are each extracted from a single cluster.

While having a different target, the work in \cite{Medeiros2007} applies clustering for the same reason as we do. The aim of that work is to discover process models that describe the sequences in an event log. A difficulty in discovering such process models is the variability in sequences. As a solution, the authors iteratively split the collection of sequences to create smaller event logs to create better models. The splitting is based on clustering to combine comparable sequences, much like our approach.

In terms of predictive process mining, this paper is part of a class of outcome prediction solutions. \cite{Tax2017} adopts LSTMs to predict the remainder (suffix) of a case by repeated next activity predictions. The same target is predicted in \cite{Taymouri2021} but then with the use of deep adversarial models. In \cite{Leontjeva2016}, the authors predict whether an active case will be compliant or not according to the business process owner. The same prediction task is executed by \cite{Francescomarino2019}, which makes use of the bucketing described above. Using techniques from text mining, \cite{Teinemaa2016} aims to early signal whether a case will have a outcome that requires intervention using unstructured textual information from events. A more detailed summary of recent outcome-oriented tasks can be found in \cite{Teinemaa2019}. Next to this, literature contains work specific to shopper behaviour prediction. Examples of these include the next interaction \cite{Hassani2021}, losing a shopper (churning) \cite{Chamberlain2017,Gunesen2021,Tariq2021}, and life-time value \cite{Chamberlain2017}. Of these, \cite{Hassani2021} also uses a process mining oriented approach, and \cite{Tariq2021} also uses neural networks for their predictions.

In this section we related \fw to existing literature in general. The literature discussed in \cref{sec:framework} focuses on alternatives to the individual steps of \cref{fig:overview:prediction,fig:overview:training}. Each step compares the choices made for \fw to alternatives in existing literature.

\section{\edititem{Framework}{}}\label{sec:framework}
In this section, we present our contributed framework \fw. We start with some notation preliminaries in \cref{sec:framework:preliminaries}. We then present a running example in predicting shopper behaviour in a supermarket as working example in \cref{sec:framework:shoppers}. We next formally introduce \fw in \crefrange{sec:framework:streaming}{sec:framework:predict}, each section formalizing its corresponding step of \cref{fig:overview:training,fig:overview:prediction}, relating it to the supermarket use case, and making a comparison to existing literature. Note that the latter comparison focuses on the design choices  and why, in the context of \fw, those are preferred over alternatives in existing work. Next, to demonstrate the generalizability of \fw, we describe another use case: paying invoices in a paint factory, in \cref{sec:framework:paintfactory}. Both use cases presented in this paper come from real-world scenarios, which are used in \cref{sec:experiments} to evaluate \fw. All symbols used in the explanation are summarized in \cref{tab:symbols}.

\begin{table*}
    \caption{Symbols used in \fw. Note that these are not parameters of the framework, just symbols indicating parts of it.}
    \label{tab:symbols}
    \centering
    \begin{tabular}{lp{5.9cm}|lp{5.9cm}}
    \toprule
    Symbol & Meaning & Symbol & Meaning \\
    \midrule
    $\entityset$ & Set of entities & $y(c,t)$ & Outcome of entity $c$ at time $t$\\
    $\entityset_t^T$ / $\entityset_t^P$ & Entities at $t$ for training / predicting & $X(c,t)$ & Feature values of entity $c$ at time $t$\\
    $\eventset$ & Set of events & $G_t^T / G_t^P$ & The clustering of $\entityset_t^T$ / $\entityset_t^P$\\
    $\eventset\downarrow^{[a,b)}$ & Events in the time frame $[a,b)$ & $\tilde{y}(g)$ & The average outcome in cluster $g$\\
    $\eventset\downarrow^{[a,b)}_c$ & Events in $\eventset\downarrow^{[a,b)}$, belonging to entity $c$ & $\tilde{X}(g)$ & Average feature values in cluster $g$\\
    $\freq(\eventset')$ & Relative frequencies of activities in $\eventset'$ &$\hat{\tilde{y}}(g)$ & Prediction for proxy entity of cluster $g$\\
    $\activityspace$ & Set of activity names  & $\hat{y}(c,t)$ & The prediction for entity $c$ at time $t$\\
    \bottomrule
    \end{tabular}
\end{table*}

\subsection{Preliminaries}
\label{sec:framework:preliminaries}
In the remainder of this paper, we use $\Naturals$ to denote all natural numbers, starting from $0$, and $\PosNaturals$ for $\Naturals \setminus \{0\}$. The positive real numbers are $\PosReals$, and $\NNegReals = \PosReals \union \{0\}$.

In this paper we consider systems that are described by a collection of \textit{events} involving \textit{entities}. Let $\eventset$ be the universe of all events and let $\entityset$ be the universe of all entities. An \textit{event} $e$ describes something that happened, labelled by activity label $e.act \in \activityspace$ (what happened?) relating to some entity $c = e.ent \in \entityset$ (to what did it happen?) at some time $e.time \in \NNegReals$ (when did it happen?). Each event has additional information $e.attributes \in \eventfeatureset$ called \textit{event attributes} and each entity $c$ has additional information $c.attributes \in \entityfeaturespace$ called \textit{entity attributes}. Let $a,b \in \NNegReals$ with $a<b$. We define $\eventset\downarrow^{[a,b)}$ as the subset of $\eventset$ with events that occur on or after $a$ and before $b$, formally:

$$\eventset\downarrow^{[a,b)} = \{e \in \eventset \vert e.time \in [a,b)\}$$

Furthermore, for an entity $c$, we define $\eventset\downarrow^{[a,b)}_c$ as the subset of $\eventset\downarrow^{[a,b)}$ with events that belong to entity $c$, formally:

$$\eventset\downarrow^{[a,b)}_c = \{e \in \eventset \vert e.time \in [a,b) \wedge e.ent = c\}$$

Let $\eventset'$ be a subset of $\eventset$. We define the function $\freq(\eventset') \in [0,1]^{\size{\activityspace}}$ as a vector that lists the relative frequency of each activity in $\activityspace$, i.e. 

$$\freq(\eventset') = (\size{\{e \in \eventset' \vert e.act = a\}} / \size{\eventset'})_{a \in \activityspace}$$

In some literature, $\freq(\eventset')$ is known as the \textit{Parikh} vector of the activities in $\eventset'$, normalized to sum to $1$.

\subsection{Running Example}\label{sec:framework:shoppers}
In this section we first describe the supermarket use case. Shoppers visit this supermarket to purchase their groceries. They may do so several times a week, each visit recording what was purchased, at what time the purchase was made, and by which shopper. The latter is known because shoppers have an incentive to use loyalty cards, as they can save up for rewards by purchasing groceries. One of the main advantages of these loyalty cards is that they allow shoppers to be identified from one visit to the next. Each of the visits can be described by a list of products purchased, including the quantity and the price paid. The collection of information in these visits over a specific time period for a single shopper constitutes the `shopper journey'.

Based on \cite{Spenrath2020}, we assign a label to each visit. The idea of this label is that it provides some information regarding the products purchased, such that visits with similar products, categories of products and quantities have the same label. These labels are learned using an unsupervised method. For full details of this labelling, please refer to \cite{Spenrath2020}. Note that \cite{Spenrath2020} is applied to \textit{single visits}; the shopper journey themselves are not categorized. This can as such be seen as part of the feature extraction that describes the shopper journey, analyzing the labels given to visits in the shopper journey. Next to this, we can also compute aggregate values for each visit, as presented in \cref{tab:receipt}.

The goal is to make a prediction about one or more future visits of a shopper. For our application this is the total spend over a given period of time, the sum of the spend in the separate visits. At the end of every week we know the total spend of the completed week, which we use to update a prediction model, and we predict the total spend of the next week, given new information about the shopper journey. To make a prediction and train the model, we extract features from several weeks of transactional data, together forming the shopper journey. In the running example we consider this to be a constant number of three weeks, but in the experimental evaluation in \cref{sec:results:shoppers} we also look at different lengths of shopper journeys for input.

Like most other real-life applications, journeys can vary wildly between different shoppers. This makes it difficult to make predictions about individual shoppers. In \fw we partition shoppers into clusters of similar shoppers. We update the prediction model using proxy shoppers averaged from the clusters instead of using individual shoppers. We further make predictions for proxy shoppers instead of for each individual shopper. In other words, we learn the behaviour of a \textit{group of shoppers} and make a prediction on their behaviour \textit{as a group}. We formalize this idea in the next sections, referring back to this use case as a running example.

In related work, different use cases are evaluated as well. In \cite{Burattin2014} the authors discover process models from an events stream describing a loan application use case. In \cite{Spenrath2019} the authors analyze an event stream relating to a maintenance service company, proposing a way to overcome gradual and recurrent concept drift in supervised learning. Another example from the process mining field is discussed in \cite{Bose2014}, where a permit application process is discussed. Use cases from supermarkets are also discussed in literature. Examples include the inter-arrival time prediction of shoppers \cite{Chen2018}, basket content predictions \cite{Doan2019a} and adoption of native shopping habits by immigrants \cite{Guidotti2021}

\subsection{Streaming setting}\label{sec:framework:streaming}
In this paper we apply machine learning in a streaming setting. In such a setting, events $e$ occur in the order defined by their time of occurrence, $e.time$. We apply batch learning to this stream: after every set amount of time we use events we have collected to update a prediction model and make predictions about entities. Without loss of generality, let $t=0$ be the time of the first event, and let the time between two batches be $1$ time unit, in other words: we update our prediction model at $t=1, 2, 3, \ldots$. After updating the model at time $t-1$, we keep collecting events up to time $t$, this collection is indicated by $\eventset\downarrow^{[t-1,t)}$. From the collection we extract information about two sets of entities: i) those that are suitable for updating the model (those for which the outcome is known) and ii) those used for prediction (those for which we want to predict the expected outcome). Depending on the use case, these two sets may overlap or be distinct.

In the supermarket use case, we collect information about transactions (events) of shoppers (entities). Each transaction describes the type of purchase ($e.act$, $\size{\activityspace}=8$), when the purchase was made ($e.time$), which shopper purchased it ($e.entity$) and additional attributes described in \cref{tab:receipt} ($\eventfeatureset = [0,1] \times \PosReals \times \PosReals \times \NNegReals \times \Naturals$). While shoppers may provide demographic information when applying for the loyalty card, this information is not part of the scope of this research ($\entityfeaturespace = \emptyset$). We update and use our model at the end of every week (one time unit is a week).

\begin{table*}
    \centering
    \caption{Values derived from a single visit.}
    \begin{tabular}{lll}
        \toprule
        Value & Description & Domain\\
        \midrule
        e.attributes.freshness & Fraction of perishable items & $[0,1]$\\
        e.attributes.item\_value & Average value of the items bought & $\PosReals$\\
        e.attributes.product\_density & Average frequency of each product & $\PosReals$\\
        e.attributes.total\_value & Total price paid for the the products & $\PosReals$\\
        e.attributes.total\_item\_count & Total number of items purchased & $\PosNaturals$\\
        \bottomrule
    \end{tabular}
    \label{tab:receipt}
\end{table*}

\subsection{Training Entities}\label{sec:framework:training}
The first step of \fw, which also applies outside of the framework, is to determine which entities, and as such which events, are used in the training phase. This is a subset of $\entityset$, which we denote as $\entityset_t^T$, with the superscript $T$ indicating that we use the set in the training phase and the subscript $t$ indicating this set is used at time $t$. The requirement for an entity to be part of $\entityset_t^T$ is that we have data needed to build the features (the input for the prediction model) as well as the outcome (the output for the prediction model). This requirement depends on the use case.

\begin{figure}[b]
    \centering
    \includegraphics[width=.45\textwidth]{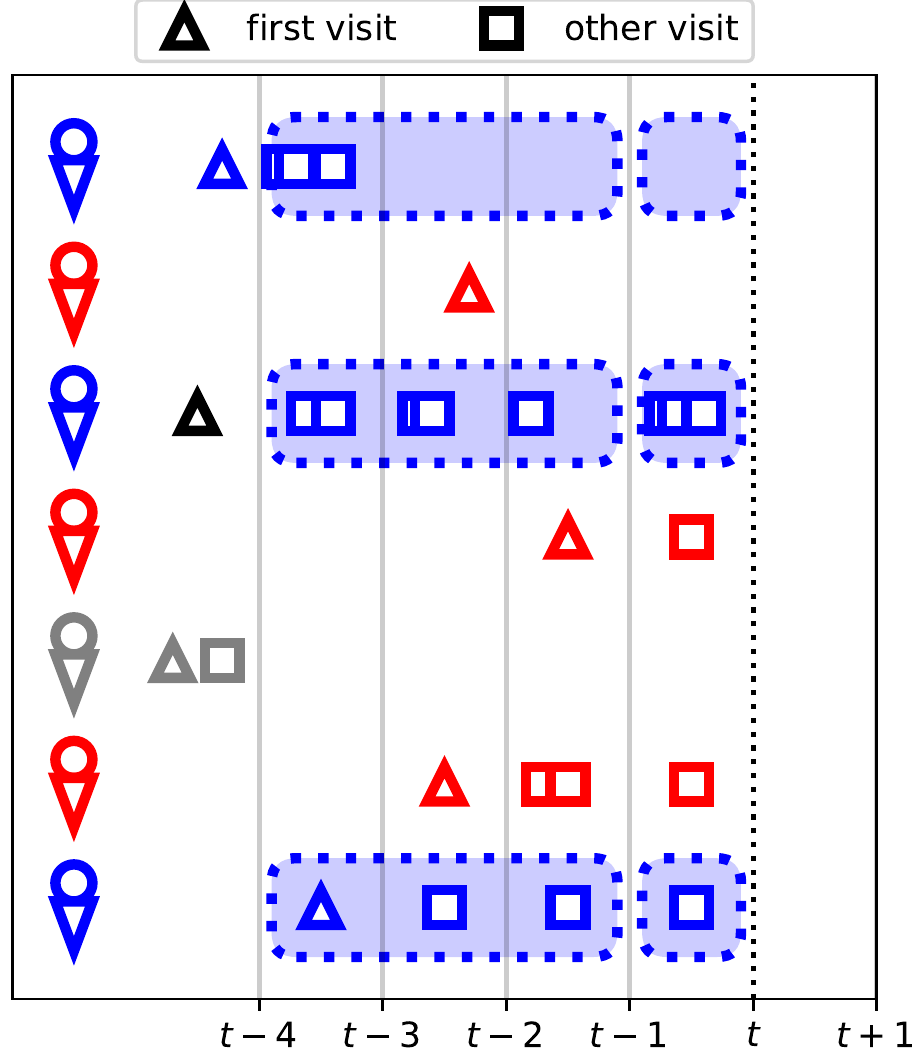}
    \caption{Shoppers selected for training. The grey shoppers did not have a visit in the three weeks from which the journey is extracted. The red shoppers do not have a sufficiently long journey since their start to be used in training. The blue shoppers, in contract, have enough data to be used in training the model.}
    \label{fig:framework:TCollection}
\end{figure}

For the supermarket use case, we require that shoppers have made transactions at the store for at least four weeks: the first three weeks are used for feature engineering to define the shopper journey, the fourth week is used to extract the outcome: the total spend of each shopper. Next to this we also require shoppers to have at least one transaction in the first three weeks to construct a shopper journey. This is visually depicted in \cref{fig:framework:TCollection}. Let $start(c)$ be the time of the first visit of shopper $c$. We formally have that

\begin{equation*}
    \entityset_t^T = \{c \in \entityset \vert start(c) < t-3 \wedge \eventset\downarrow^{[t-4,t-1)}_c \neq \emptyset\}
\end{equation*}

In the setting of \fw we limit ourselves to use cases where it is clear to which entity an event belongs. Put differently, each entity can be considered as a `generating process' that emits events related to it, and we can distinguish different sources. This is a common in many real-world datasets, specifically to process mining \cite{VanderAalst2016} (events have a `case identifier'), retailer data \cite{Chen2018,Doan2019a,Ariannezhad2021} (visits belong to some sort of loyalty card) and online customer journeys \cite{Bai2019} (journeys of web clicks that belong to a specific user). As such, we argue that this is a valid assumption. There is related work that tries to identify the entity to which new events belong. In \cite{Unnikrishnan2020}, the authors predict the generating sources by linking incoming information to entities by finding resemblance between new and existing data. In \cite{Ferriera2009}, the authors also deal with events for which the entities (case identifiers) are not known, solving the problem by estimating a process model from incoming events and computing the probability that it belongs to an existing entity. A similar problem is solved in \cite{Bayomie2016}. Note that these works are solving the problem of relating events to entities, which is different from the problem \fw solves.

\subsection{Encoding}\label{sec:framework:encode}
For encoding an entity we combine information from past events of that entity. Specifically, we create values $X(c,t)$ and $y(c,t)$. The former describes the input data for the prediction model, the latter describes the outcome for that entity.

In the supermarket example, at time $t$ we use the past week ($[t-1,t)$) for the outcome (the total spend of the shopper). The outcome is simply the sum of the value of all transactions:

\begin{equation*}
    y(c,t) = \sum_{e \in \eventset\downarrow^{[t-1,t)}_c} e.attributes.total\_value
\end{equation*}

With the help of domain knowledge\footnote{The authors gratefully thank the company [\textit{redacted}] for making their data available for this project and their useful feedback on \fw.} we convert events of a single shopper to an encoding of the shopper journey. To this end, we compute descriptive statistics of the transactions a shopper does in a single week, as such creating three sets of descriptive values to define the journey of a shopper over the period of three weeks. We take the average of the first three values of \cref{tab:receipt} and the sum for the final two values. We use the sum for e.attributes.total\_value and e.attributes.total\_item\_count as this helps distinguishing shoppers with a large total spend from shoppers with a smaller total spend. Next to this we also use the relative frequency of each label and the number of visits. In total this creates a matrix of 14 rows (5 aggregates, 1 for the number of visits, and 8 activity label frequencies) and 3 columns (1 for each week).

In related work, there are alternatives to encoding, which is sometimes also referred to as `embedding'. In other works of the process mining field (where entities are often referred to as \textit{cases}), the most frequently used method is to limit the number of events considered (prefix) per entity to create evenly sized embedding tensors. Each row describes a single event, and each column describing a dimension of the event. These are dimensions such as a one-hot-encoded activity label or any of the event or entity attributes. Examples of such uses include \cite{Francescomarino2019,difrancescomarino2018,Leontjeva2016}. The disadvantage of this approach in the supermarket use case is that the number of events in a time frame is highly relevant. The solution where all events in a single week are aggregated together allows the use of all events (including information about the number of events), while having evenly-shaped vectors for each entity (shopper) to accommodate machine learning model inputs. The frequency-based feature values we use are also used in for example \cite{Song2009}. Further alternatives are Hidden Markov Models \cite{Leontjeva2016}, and frequencies of common subsequences \cite{Medeiros2007,bose2012}. The disadvantage of these is that finding such models and relevant subsequences can be computationally expensive. Finally, the work of \cite{Chamberlain2017} further suggests the use of automatically learned features over handcrafted ones for the prediction.

\subsection{Clustering}\label{sec:framework:cluster}
Traditionally, we would update a prediction model using the predictor-outcome values ($X(c,t), y(c,t)$) from \cref{sec:framework:encode} for the entities $c$ in $\entityset_t^T$. In this paper we propose an intermediate step. Instead of using \textit{individual} entities to train the model, we first create \textit{clusters} of entities, and then update the model with proxy entities of each cluster. The number of clusters we use depends on the size of $\entityset_t^T$. Let $\rho \in \PosNaturals$, the number of clusters $\entityset_t^T$ is partitioned into is:

\begin{equation}
    \label{eq:rho}
    k = \ceil{\frac{\size{\entityset_t^T}}{{\rho}}}
\end{equation}

\noindent Put differently, we create a number of clusters such that the average cluster size is roughly equal to $\rho$. This leaves us with a group of $k$ clusters, which is denoted as $G_t^T$, the super- and subscript defined similar as in $\entityset_t^T$. The advantage of choosing the average cluster size as parameter instead of the number of clusters is that the number of entities from one time step to the next may differ. As a result, the number of clusters learned will also differ. As we update the prediction model with a proxy entity from each cluster, this means that if a time step has more entities suitable for updating prediction model, the prediction model is updated with more proxy entities. Conversely, for a smaller $\entityset_t^T$ the prediction model is updated with fewer proxy entities. As we discuss in \cref{sec:framework:paintfactory}, the number of entities suitable for prediction varies greatly over time in the paint factory use case. In the original work, we used the number of clusters as the parameter instead, which influenced the stability of the results for the paint factory use case.

$\rho$ controls how many clusters are computed in the clustering step of the training (and prediction) phase. This introduces a trade-off. For lower values of $\rho$, fewer entities are clustered together, meaning that each proxy entity is closer to the original entities, the predictions are closer to an individual entity level, as such making them more useful. However, increasing $\rho$ removes the effect outliers have on the predictions and training. Apart from this, an increase in $\rho$ also decreases the number of proxy entities, which on the one hand potentially reduces number of training points and hence the quality of the model.

Of special note are two values: $\rho=1$ and $\rho=\size{\entityset}$. For $\rho=1$, \fw is effectively equal to not using it: proxy entities are the same as actual entities and individual entities are used for training (and, later on, predicting). As such, the value $\rho=1$ can be considered as a comparison method, where \fw is not used. For $\rho=\size{\entityset}$, all entities are added to a single cluster; only a single proxy entity is used in prediction and training. These extreme values match the trade-off descriptions given earlier.

\fw does not depend on the choice of the clustering algorithm or the distance metric used for the clustering itself, just that there is \textit{a} way of properly separating the entities into a given number of subsets. This is mostly because the availability of clustering algorithms and distance metrics also depends on the use case. For instance, the supermarket use case requires a different approach than the paint factory use case. \fw focuses on improving the predictive quality of existing models by grouping together similar entities, independent of the way in which these groups are created.


For the supermarket use case, we measure the distance between two shoppers by the difference in the progression of their journey over the three weeks. More specifically, for each of the 14 features discussed in \cref{sec:framework:encode} we fit a linear function for each shopper on the values of that feature over time (in the journey), estimating feature $f_v$ as $a_v \cdot t + b_v$ with residual $r_v$. Shoppers with similar values for these coefficients have a similar average ($b_v$), progression over time ($a_v$) and linear fitness ($r_v$). Using these $3 \cdot 14 = 42$ coefficients, we cluster the shoppers using the Euclidean distance and Lloyd's algorithm for $k$-medoids \cite{Lloyd1982}. Because we are clustering a large number of shoppers at every time step, we use an efficient approximation of the Euclidean distance, which also requires the use of $k$-medoids over $k$-means. Details of this approximation are discussed in \cite{Spenrath2022}.

Specifically for the supermarket scenario, we have chosen the measure the similarity of the linear fits of the individual shoppers as (inverse) distance metric for the clustering. An alternative to this approach is discussed in \cite{VandenBerg2021}. The authors propose a iterative learning approach that learns clusters of entities and a distance metric at the same, also with event-based features. The technique incorporates domain knowledge. This technique can be used in a preprocessing phase of \fw to determine a relevant distance metric over the entities. \cite{Medeiros2007} uses the Euclidean distance on their defined encoding, together with $k$-means. In \cite{Francescomarino2019}, DBSCAN \cite{Ester1996} is used for the clustering of sequences of events. DBSCAN is not directly applicable to \fw as we define the number of cluster we aim for by design, where DBSCAN is based on different parameters that controls the density of the clusters instead of the number of clusters.

\subsection{Averaging}\label{sec:framework:average}
Having found the partition $G_t^T$ of entities, we next compute the proxy entities. For cluster $g \in G_t^T$, this is done by taking the average of $X(c,t)$ for each entity in $g$, creating proxy entity $\tilde{g} = (\tilde{X}(g),\tilde{y}(g))$ with proxy feature values $\tilde{X}(g)$ and proxy outcome $\tilde{y}(g)$. As such, we create a total of $k = \ceil{\frac{\size{\entityset_t^T}}{{\rho}}}$ proxy entities. Note that this assumes that each dimension in $X$ can be averaged.

In the supermarket use case, each proxy entity $\tilde{g}$ is again represented as a $14$ by $3$ matrix, the value at row $a$ and column $b$ containing the average feature value at row $a$ and column $b$ over all shoppers in $g$. This creates a total of $k = \ceil{\frac{\size{\entityset_t^T}}{{\rho}}}$ proxy shoppers. Each of these features is a numerical feature, with a meaningful average if aggregated over multiple shoppers.

\begin{figure}
    \centering
    \includegraphics[width=.45\textwidth]{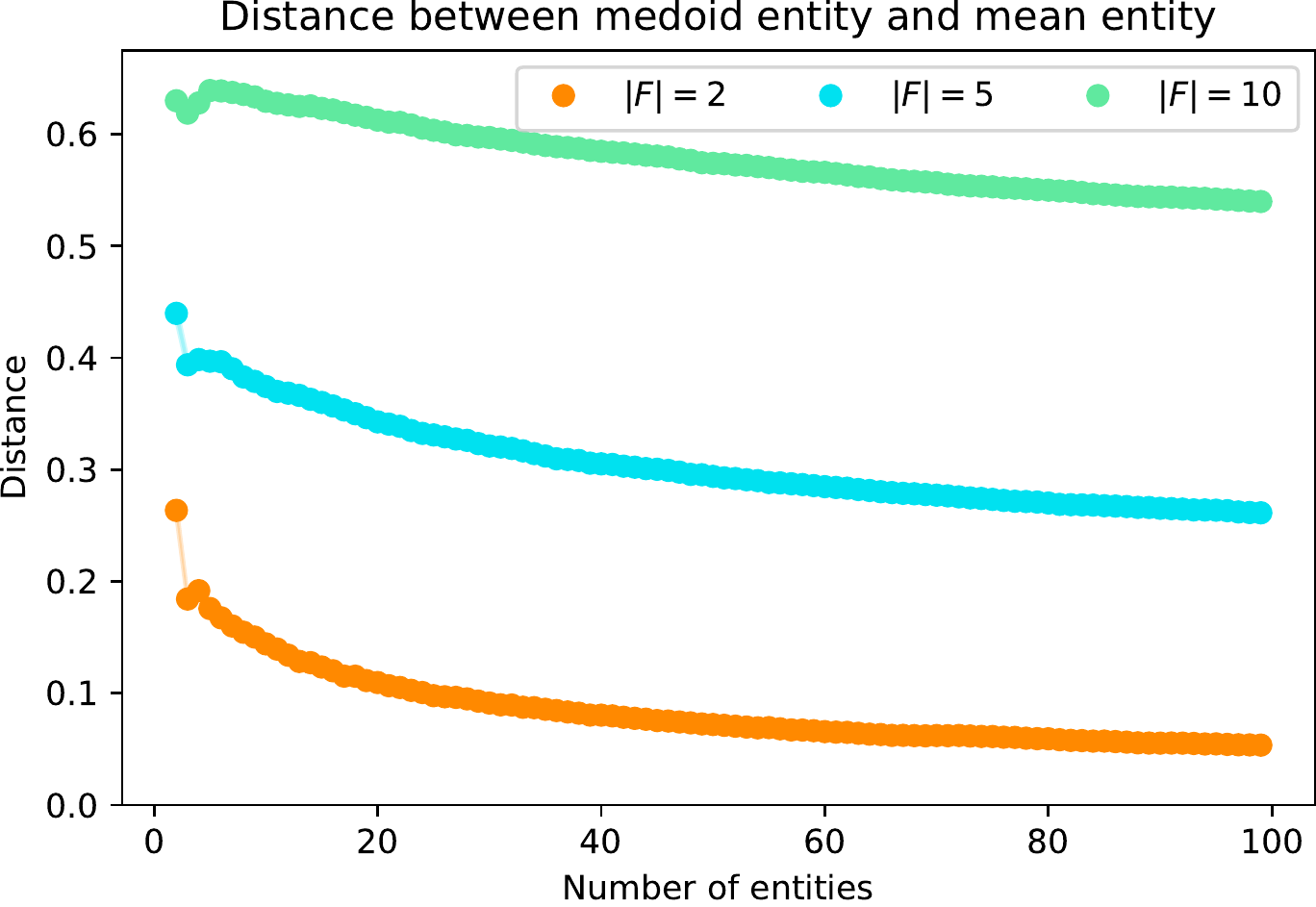}
    \caption{Average Euclidean distance between mean and medoid entity, taken over $1000$ random samples as function of the number of entities, for $2$, $5$ and $10$ dimensions.}
    \label{fig:framework:meanmedoid}
\end{figure}

In \fw we take an average over a cluster of entities. Another options is to use one of the entities as representative. This can either be the medoid, which is a side-product of $k$-medoids, or the entity closest to the mean; though these may be the same. The latter is used in \cite{Lin2017} to sample the majority class of an unbalanced dataset in supervised machine learning. Depending on the set of entities, the mean entity may be very close to the medoid entity. As shown in \cref{fig:framework:meanmedoid}, the distance between these two decreases when the number of entities over which they are taken increases. As a result, the medoid entity is less representative of the mean entity for a lower number of entities than for a higher number of entities. This may influence the results between different values of $\rho$, which have precisely this difference. As such, we use the mean entity as the proxy entity.

\subsection{Updating}\label{sec:framework:update}
In the final part of the training phase we train/update the model. In the first time step of the streaming setting a model is created from scratch (cold start), and in subsequent time steps the model is updated. As such the only requirement that we have for the model is that it is incremental: after initially training, updates to it with new training data must be possible.

\fw does not depend on the choice of the prediction model, its outcome type (classification or regression) or any of its hyper-parameters, just that there is \textit{a} prediction model that can be updated after initial training. This is mostly because the most suitable model depends on the use case. For instance, the supermarket use case requires a different approach than the paint factory use case. \fw focuses on improving the predictive quality of existing models by grouping together similar entities, independent of the model.

In the supermarket use case, we use a long-short term memory RNN (LSTM). LSTMs are trained on \textit{sequences} of data and allow for incremental training. This makes them ideal for this use case. In an online training setting, we train the model from scratch in the first time step, and update it at every next time step. LSTMs are trained on sequences of data. Each element in the sequence is one week of shopper journey, so a single column of $\tilde{X}$. Columns are then added to the LSTM one at a time, which is trained to predict $\tilde{y}$.

In related work on predicting based on sequences of data, LSTMs are more often used. \cite{Tax2017} uses it to predict the remainder (suffix) of a case by repeated next activity predictions. \cite{Taymouri2021} uses deep adversarial models. Next to deep learning models, a decision tree or random forest is often used for predictions based on events, such as in \cite{Leontjeva2016,Francescomarino2019,Teinemaa2016}. The encoding in these works does not result in sequences, making the use of a decision tree/random forest more applicable than in the supermarket scenario. Next to this, decision trees and random forests are not incremental by themselves. Incremental variants do exist as Options Trees and their extensions, discussed in \cite{Garcia2006,Bifet2009} and implemented in \cite{skmultiflow,SOFTWARE:MOA}. These have also been used in event stream predictions such as \cite{Spenrath2020MT,Li2019}. We use one of these incremental variants in the paint factory use case of \cref{sec:framework:paintfactory}.

\subsection{Prediction phase}\label{sec:framework:predict}
In the prediction phase (\cref{fig:overview:prediction}) we apply similar steps as in the training phase, with some minor differences.

\textbf{Prediction Entity Collection}: We first determine which entities are suitable for making a prediction for. We denote this subset of $\entityset$ as $\entityset_t^P$, the superscript now indicating that we use the subset in the predictions at time $t$. The requirement for an entity to be part of $\entityset_t^P$ is slightly different from $\entityset_t^T$: we do not require the outcome of an entity, just the data needed to build the features used by the prediction model. The exact requirement again depends on the use case.

\begin{figure}[b]
    \centering
    \includegraphics[width=.45\textwidth]{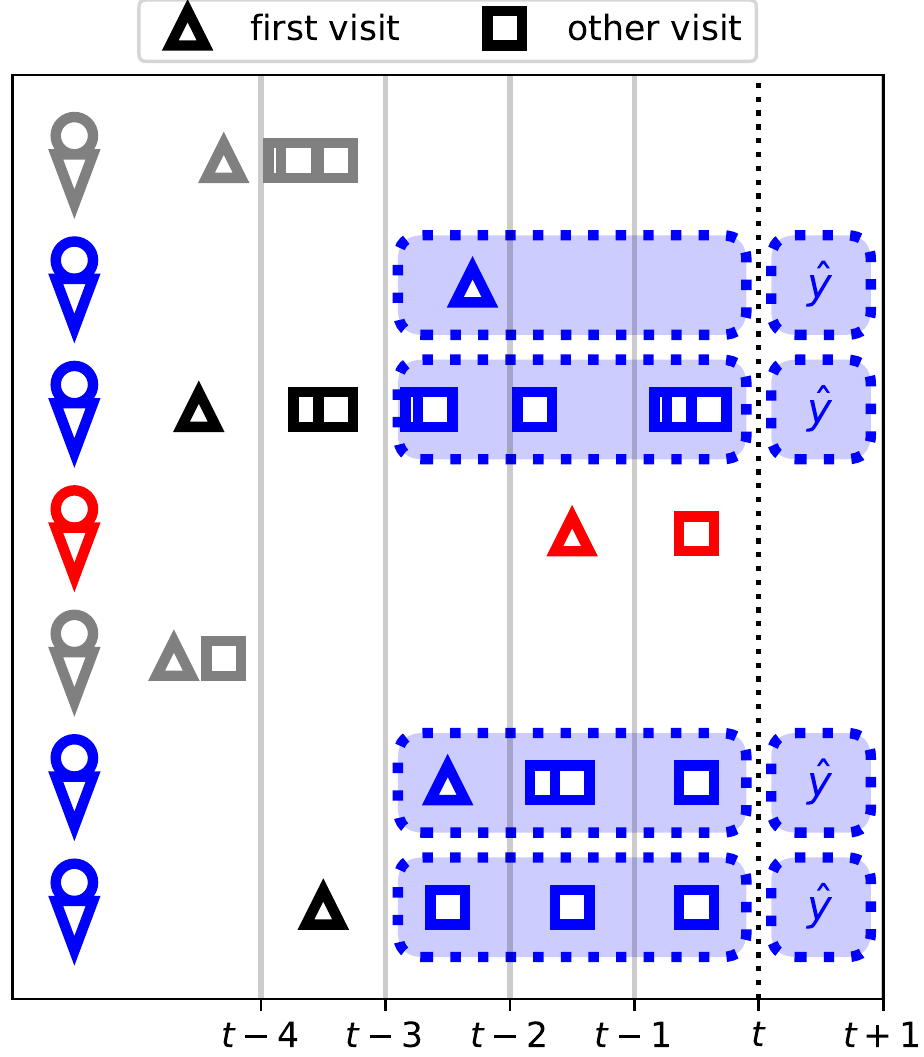}
    \caption{Shoppers selected for prediction. The grey shoppers did not have visits to construct a journey in the past three weeks. The red shoppers did have visits, but only started in the past two weeks and are therefore also not used for prediction. The blue shoppers have transactions in the last three weeks and started their visits in time, they are used for prediction.}
    \label{fig:framework:PCollection}
\end{figure}

In the supermarket setting, we require that shoppers have made transactions at the store for at least three weeks ($[t-3, t)$); these together form a shopper journey that can be encoded to be used in the clustering (to find similar shoppers) and later in prediction (after aggregating shopper clusters to a proxy shopper). This is in contrast to the training collection, where four weeks were needed to also have the outcome value. The outcome value for this data is from the week following $t$ ($[t, t+1)$), the value we want to predict. Next to this we also require shoppers to have at least one transaction in those weeks (so shoppers that had their last purchase more than three weeks ago are not used in the prediction phase). This is visually depicted in \cref{fig:framework:PCollection}. As such we have:

\begin{equation*}
    \entityset_t^P = \{c \in \entityset \vert start(c) < t-2 \wedge \eventset\downarrow^{[t-3,t)}_c \neq \emptyset\}
\end{equation*}

\textbf{Encoding, Clustering, Averaging}: For the encoding in the prediction phase we apply the same procedure as for the training phase, with the difference being the entities on which the steps are applied: $\entityset_t^T$ instead of $\entityset_t^P$. This creates partition $G_t^P$, in which each cluster $g$ is averaged into a proxy entity with feature values $\tilde{X(g)}$.

For the supermarket, all three of the encoding, clustering and averaging steps are the same, but now applied to $\entityset_t^P$ instead of $\entityset_t^T$.

Note that because of the requirements for $\entityset_t^P$ and $\entityset_t^T$, we have that if a shopper is in the shopper collection for predictions at time $t$ ($\entityset_t^P$) it will also be in shopper collection for training at time $t+1$ ($\entityset_{t+1}^T$), by which time we have the actual outcome as well. In other words, $\entityset_t^P = \entityset_{t+1}^T$ for all $t$. As such, the encoding and clustering only needs to happen once every time step, we can reuse the results of the prediction phase of the last time step for the prediction phase in this time step.

\textbf{Prediction}: In the prediction step, we use the proxy entity feature data, $\tilde{X}(g)$ to predict the outcome $\hat{\tilde{y}}(g)$ for the proxy entity of cluster $g$. We then also assign this same prediction to all entities in $g$. In the next time step we have the ground truth outcome for each of the entities; we can then assess the quality of the model at time $t$ by comparison of the ground truth with the predicted outcome. The metrics used for this step depend on the use case.

For the supermarket use case we compute four metrics to assess \fw. These metrics are either focused on accuracy or on usefulness of the prediction, to asses the trade-off show in \cref{fig:problem}. The metrics are summarized in \cref{tab:framework:metricsshopper}.

\begin{table*}[]
    \centering
    \caption{Metric used to asses \fw in the supermarket use case}
    \label{tab:framework:metricsshopper}
    \begin{tabular}{lp{10cm}l}
         \toprule
         Metric & Explanation & Focus\\
         \midrule
         Cluster $\mathbf{RMSE}$ & Comparison of proxy prediction $\hat{\tilde{y}}(g)$ and ground truth $\tilde{y}(g)$ & Accuracy\\
         Shopper $\mathbf{RMSE}$ & Comparison of shopper prediction $\hat{y}(c,t)$ and ground truth $y(c,t)$ & Both\\
         $\mathbf{F_1}$ & Ability to find shoppers with the highest turnover drop & Usefulness\\
         $\mathbf{APE}$ & Ability to predict the total turnover & Usefulness\\
         \bottomrule
    \end{tabular}
\end{table*}

The first is measuring the root mean square error $\mathbf{RMSE}$ between the predicted turnover of a cluster ($\hat{\tilde{y}}(g)$) and the actual turnover of the cluster ($\tilde{y}(g)$), taken over all clusters in $G_t^P$. This says something about the accuracy of the prediction. The second measure is similar, but now comparing the predicted value of a shopper ($\tilde{y}(c,t)$) and the ground truth of the shopper ($y(c,t+1)$), taken over all shoppers in $\entityset_t^P$. This says something about the accuracy and the usefulness of the prediction. We distinguish between these two metrics as cluster-$\mathbf{RMSE}$ and shopper-$\mathbf{RMSE}$ respectively. The third use is a downstream prediction task, predicting which shoppers have the highest loss in spend from one week to the next. We compare the spend of this week ($y(c,t)$) with the spend of last week ($y(c,t-1)$) for all shoppers, and label the decile with the highest decrease in spend with \texttt{TRUE}, and all others with \texttt{FALSE}. This can be used by the retailer to apply targeted marketing: only reaching the shoppers that are the most interesting as they decrease their spend by the largest amount. We also determine predicted labels based on the predicted spend for the subsequent week and use these with the ground truth labels to compute the $\mathbf{F_1}$ score. This metric focuses on usefulness. The fourth use is in the prediction of the total turnover of the next week, computed by summing the individual (predicted) spend values over all shoppers. This results in a single value for the ground truth ($T$) and a single value for the prediction $\hat{T}$, from which we compute the absolute percentage error $\mathbf{APE}$, by dividing the difference by the ground truth ($\smallsize{T-\hat{T}}/T \cdot 100\%$). This metric also focuses on usefulness.

\subsection{Paint Factory}\label{sec:framework:paintfactory}
\begin{sidewaysfigure}
    \centering
    \includegraphics[width=0.9\textwidth]{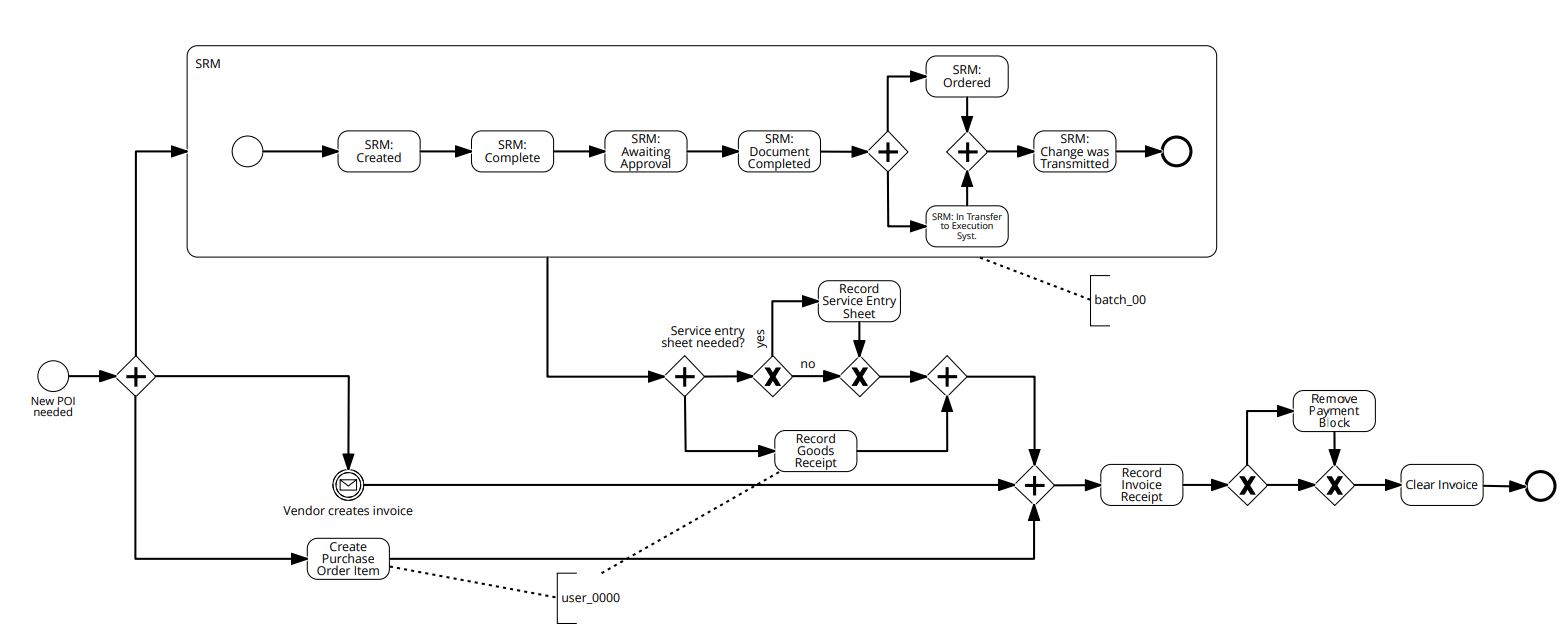}
    \caption{Process overview as constructed by \cite{bpic2019winners}}
    \label{fig:processoverview}
\end{sidewaysfigure}

Having discussed \fw in general, as well as the application within the supermarket use case, we discuss how it can be applied in a different use case. This use case concerns the purchasing process of a paint factory. Each purchase comes with an invoice that is at some point in time received and at some later point in time paid. We are interested in predicting how much time there will be between receiving and paying an invoice. An overview of the purchasing process is shown in \cref{fig:processoverview}. To this end, we collect information on every step in the process.

\textbf{Streaming Setting:} In this use case we deal with invoices as the entities. Whenever one of the process steps of \cref{fig:processoverview} occurs, and event $e$ is saved. This event details when it occurred ($e.time$), to what invoice it belongs ($e.entity$), and what happened ($e.act$). Two activity labels are of particular interest: `Vendor Creates Invoice' (\texttt{VCI}) and `Record Invoice Receipt' (\texttt{RIR}) which respectively denote the creation and payment of the invoice that belongs to an order. Other than the time, entity and activity, no additional information is recorded on events ($\eventfeatureset = \emptyset$). Each invoice does contain additional information, presented in \cref{tab:orderfeatures}. For more details on this dataset see \cite{bpic2019}. We update and use the prediction model at the end of every day (one time unit is a day).

\textbf{Training Entities:} For the set of invoices that can be used for training at step $t$ we require that the invoice has been paid. This is noted by the occurrence of an event with activity label \texttt{RIR} for that invoice. Let $rir(c)$ be the timestamp of that event. We use the invoices for which this event occurred since $t-1$ for updating the prediction model:

\begin{equation*}
    \entityset_t^T = \{c\in\entityset \vert rir(c) \in [t-1,t)\}
\end{equation*}

\begin{figure}
    \centering
    \includegraphics[width=.45\textwidth]{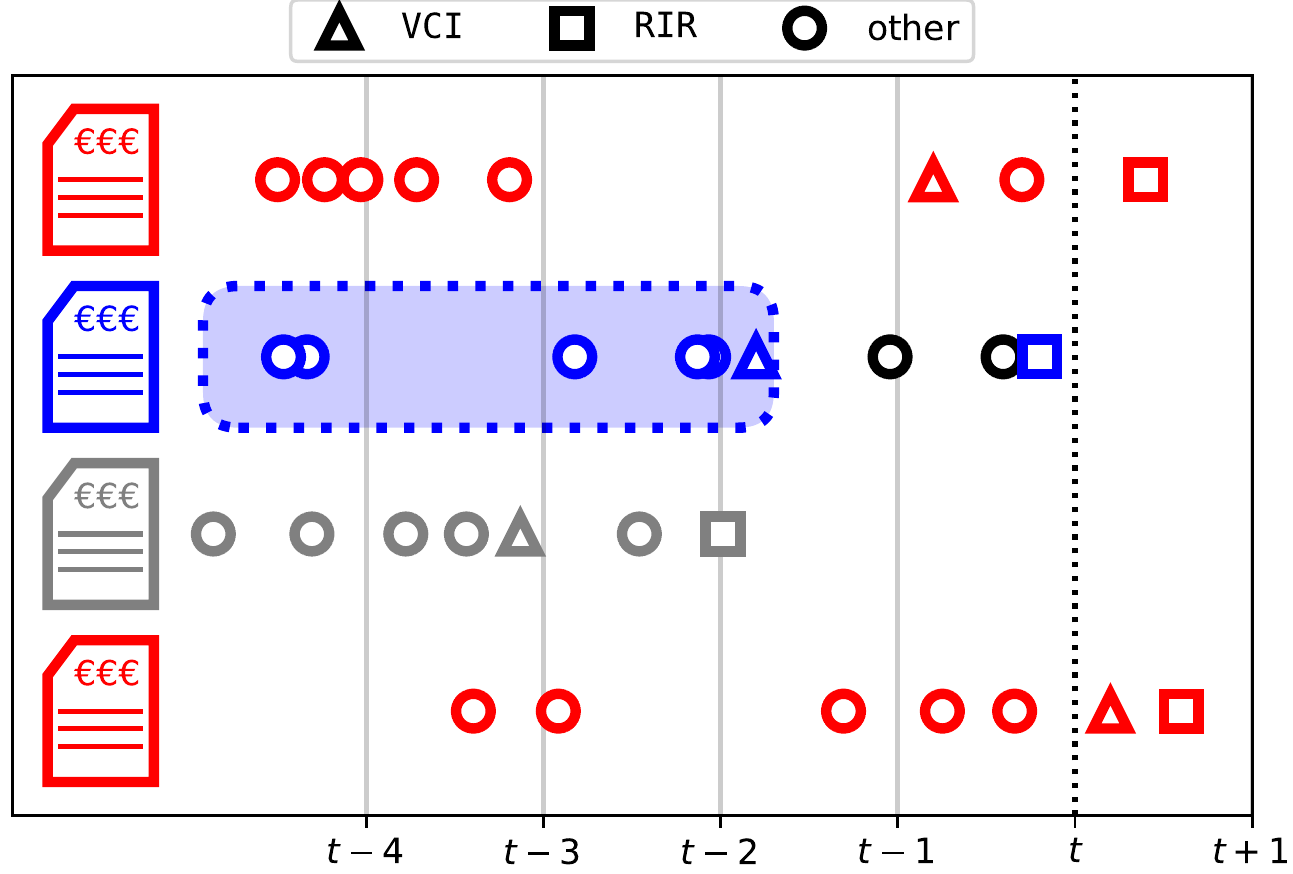}
    \caption{Invoices selected for training. The invoices in grey have already been used for training. Invoices in red did not get \texttt{RIR}, so we do not know the duration yet and as such cannot use these invoices for training. Invoice in blue did get \texttt{RIR} since the last training time, so we can use them for training.}
    \label{fig:framework:TCollectionPF}
\end{figure}

This is visualized in \cref{fig:framework:TCollectionPF}.

\textbf{Encoding:} the outcome of an invoice is defined as the time difference between the payment of the invoice (when \texttt{RIR} occurs) and the sending of the invoice (when \texttt{VCI}) occurs. Let $vci(c)$ be the time of the latter. We have that $\hat{y}(c,t) = rir(c)-vci(c)$. For the prediction model, it is important to realize that we can only use information available at the moment of prediction. That is, for the construction of prediction features of invoice $c$ we should only use data available by $vci(c)$, even if we also have information from the period $[vci(c), rir(c))$ when we update our model. The predictor values come from two sources: the relative frequency of each activity and the invoice attributes. The latter of these are the ones described in \cref{tab:orderfeatures}. For the relative frequencies we only consider events that occur up to the prediction point, i.e. $\freq(\eventset\downarrow^{[0,vci(c))}_c)$. As such, $X(c,t) \in [0,1]^{\size{\activityspace}} \times  \entityfeaturespace$.

\textbf{Clustering:} Some of the features of $X(c,t)$ are categorical and some are numerical. For the clustering we apply Gower's distance \cite{Gower1971} to deal with this, again applying $k$-medoids, as $k$-means does not work for categorical values.

\textbf{Averaging:} We apply one-hot-encoding to the categorical features to allow for averaging them for the proxy invoice encoding.

\textbf{Updating:} As we are not dealing with sequences of data, we do not require the use of models that can handle them. To further validate \fw we therefore use a different model, the Adaptive Hoeffding Option Tree \cite{Bifet2009}. This is an incremental model, we can update after initial training.

\textbf{Predicting:} For the set of invoices that can be used for prediction at step $t$ we require that the invoice is received. This is noted by the occurrence of an event with activity label \texttt{VCI} for that invoice:

\begin{equation*}
    \entityset_t^P = \{c\in\entityset \vert vci(c) \in [t-1,t)\}
\end{equation*}

\begin{figure}
    \centering
    \includegraphics[width=.45\textwidth]{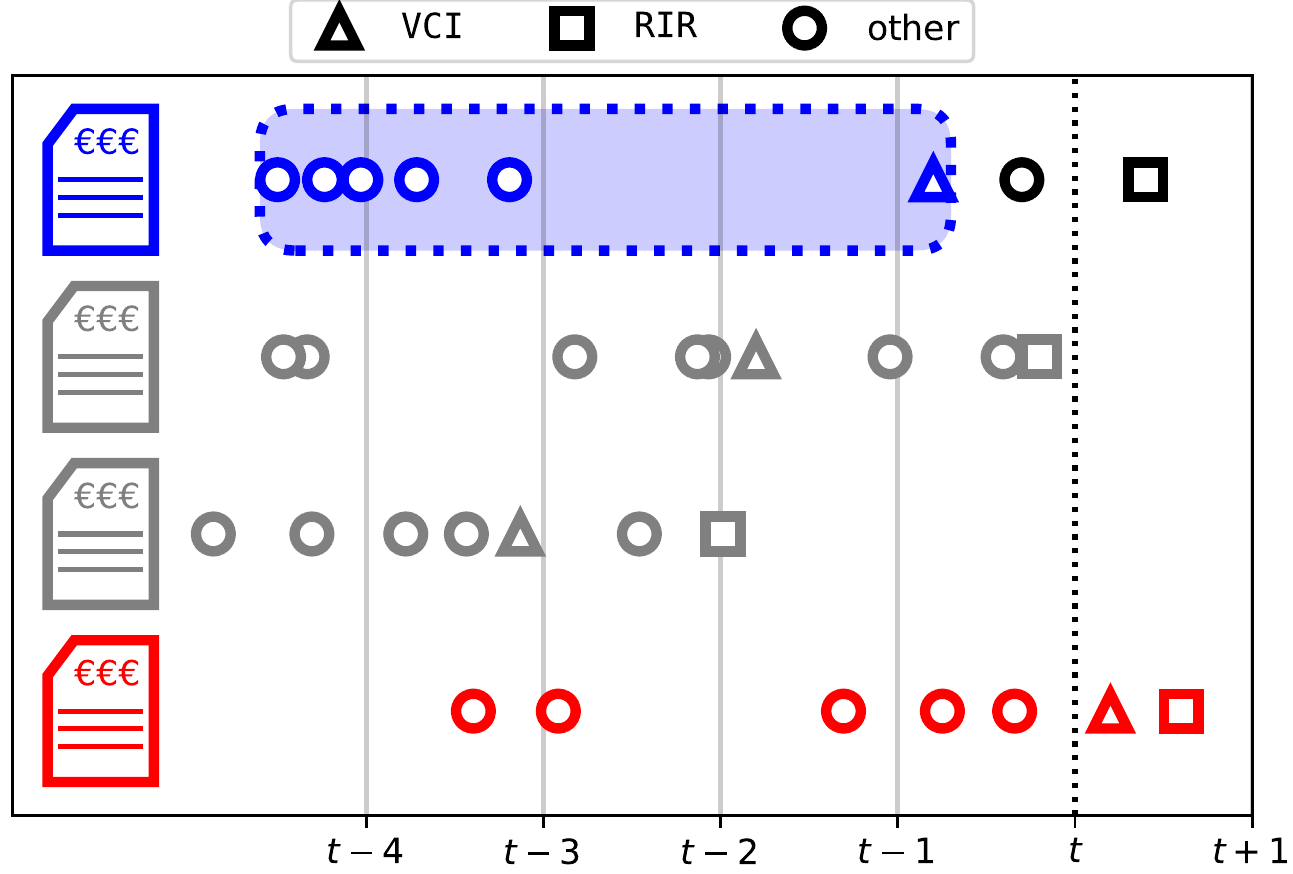}
    \caption{Invoices selected for predicting. The invoices in grey have already had their prediction. Invoices in red did not get \texttt{VCI}, so we can not make a prediction yet. Invoices in blue did get \texttt{VCI} since the last prediction time, so we can make a prediction for them.}
    \label{fig:framework:PCollectionPF}
\end{figure}

This is visualized in \cref{fig:framework:PCollectionPF}.

As a consequence of the definition of our prediction, we make exactly one prediction for each invoice, and we use each invoice exactly once to update the model. Put differently, for $t \neq t'$ we have $\entityset_t^P \cap \entityset_{t'}^P = \emptyset$ and $\entityset_t^T \cap \entityset_{t'}^T = \emptyset$. This is in contrast to the supermarket use case where shoppers are used for training and prediction at the end of every week. This makes sense from a model perspective as well since the model is trained using feature data of invoices \textit{up to} the \texttt{VCI} activity. Adding more data after that would change the state of the (proxy) invoice with data the model is not trained on. This is different from the supermarket use case where shoppers shop continuously instead of a process that progresses and eventually ends such as for the paint factory. For consistency, we do still use the notations for $X$ and $y$ that involve $t$.

As metric for evaluation we use the $\mathbf{RMSE}$ between the actual ($y(c,t)$) and predicted ($\hat{y}(c,t)$) duration of paying the invoice, in other words, the $\mathbf{RMSE}$ on the entities. 

\begin{table*}[t!]
    \centering
    \caption{Invoice features used in the paint factory use case.}
    \label{tab:orderfeatures}
    \begin{tabular}{lll}
    \toprule
    Feature & Explanation & Values\\
    \midrule
    Company & Vendor from which the product is ordered & Categorical (3)\\
    Document Type & Type of order & Categorical (3)\\
    GR-Based Inv. Verif. & Flag regarding requirements of the process & Boolean\\
    Goods Receipt & Flag regarding requirements of the process & Boolean\\
    Item Category & Further purchase/invoice regulations & Categorical (3)\\
    Item Type & Type of product purchased & Categorical (5)\\
    Spend area text & Department for which the order is created & Categorical (21)\\
    Spend classification text & Additional information to Spend area text & Categorical (4)\\
    \bottomrule
    \end{tabular}
\end{table*}

\begin{table}[]
    \centering
    \caption{Summary of the use cases to which \fw is applied.}
    \label{tab:framework:summary}
    \begin{tabular}{lll}
         \toprule
         Definition & Supermarket & Paint Factory\\
         \midrule
         Event & Visit & Process Step \\
         Entity & Shopper & Invoice \\
         Time unit & 1 week & 1 day \\
         Model Input& Shopper journey & Process execution\\
         Entity outcome & Weekly spend & Payment duration \\
         \bottomrule
    \end{tabular}
\end{table}

\subsection{Discussion}
As discussed above, \fw can be used for different use cases. The two discussed are summarized in \cref{tab:framework:summary}. In this section we review \fw and discuss important aspects in the design.

As mentioned, \fw is about the use of clustering to improve the effectiveness of prediction models. As such, the clustering method and the prediction model used in \fw are not a design choice, but depend on the use case \fw is applied to. In principle, any method that partitions a set of entities in a specified number of subsets can be used for this step. In the above we have discussed the use of the approximate Euclidean distance from \cite{Spenrath2022} and the use of Gower's distance from \cite{Gower1971} together with $k$-medoids. In the experimental evaluation we also compare the results when using $k$-medoids with those using randomly assigning entities to clusters. The same holds for the prediction model where in principle any learner that can be updated after initial training can be used in \fw. In the above we have discussed the use of an LSTM and the Adaptive Hoeffding Option Tree, but even static learners that are completely retrained at every time step are a valid model. \fw is about the combination of \textit{a} clustering method and \textit{an} incremental learning mechanism to improve the quality of the predictions. Existing literature focuses on which prediction models are used \cite{skmultiflow,Teinemaa2019,Tariq2021} and how entities are clustered instead \cite{Lin2017,VandenBerg2021,Leontjeva2016}; in that sense, the contribution is orthogonal to these. This holds for prediction tasks in general, and also specific to recommendation systems and other related work on shopper behaviour prediction.

This paper and in particular the experimental evaluation discusses the effect \fw has on the quality and usefulness of the predictions. \fw may also have an influence on the computation time. On the one hand, the clustering increases the duration of the training/testing phase. On the other hand, the use of proxy entities means that the model is trained on fewer datapoints and makes predictions for fewer datapoints. These two changes in the training and prediction phase have opposing effects on the computation time. It is possible that the added time required for clustering is outweighed by the reduced time required for training/predicting. However, an analysis of this is beyond the scope of the current paper.

\section{Experimental Evaluation}\label{sec:experiments}
In this section we discuss the experimental evaluation of the method described in \cref{sec:framework}. We first discuss the supermarket use case from \cref{sec:framework:shoppers} in \cref{sec:results:shoppers}, followed by the results from the paint factory use case from \cref{sec:framework:paintfactory} in \cref{sec:results:paintfactory}. We then conclude with lessons learned from both datasets in \cref{sec:results:conclusion}. The implementation of \fw as well as the experiments can be found at \url{https://anonymous.4open.science/r/ITEM-CAPiES}.

\subsection{Supermarket Shopper Experiments}\label{sec:results:shoppers}
For the experiment, we use data from a real-life supermarket. The visit data comes from $39$ weeks and contains over $160 000$ shoppers with at least one purchase in that time. The number of visits per shopper varies between $1$ and $312$ with a mean of $55.0$ and a standard deviation of $21.8$. Because of privacy reasons, this dataset cannot be made publicly available. We discuss the experimental setup in \cref{sec:eeshopper:setup} and discuss the results in \cref{sec:eeshopper:taurmse} through \ref{sec:eeshopper:rhof1time}.

\subsubsection{Experimental Setup}\label{sec:eeshopper:setup}
\edititem{For the experiment we vary the average number of shoppers per cluster $\rho$, taking values $2^1=2, 2^3=8, 2^5=32, \ldots, 2^{17}$ (i.e. in multiples of $4$), the last of these being just smaller than the largest value of $\size{\entityset_t^P}$. We also add the special values of $\rho=2^0=1$ and $\rho=\size{\entityset}$ as competitors that do not apply clustering. In the running example we used a shopper journey of three weeks to make predictions. We also vary this value, indicated by the parameter $\tau$, exploring values $2, 3, 4, \ldots, 9$ weeks. Note that $\tau$ is a variable for the supermarket use case only; it is not part of the framework, nor does the paint factory use case use it. As such we test \fw on $11 \cdot 8$ combinations of $\rho$ and $\tau$.}{For the experiments, we vary the sequence $\tau$ and the average number of shoppers per cluster $\rho$. For $\tau$ we take $2, 3, \ldots, 9$. For $\rho$ we take values $2^1=2, 2^3=8, 2^5=32, \ldots, 2^{17}$ (i.e. in multiples of $4$), the last of these being just smaller than the largest  value of $\size{\entityset_t^P}$. We also add the special values $\rho=2^0=1$ and $\rho=\size{\entityset}$ as competitors not using any clustering. In total, we conduct $8 \cdot 11 = 88$ experiments.}

For the LSTM architecture we use the one defined by \cite{Tax2017}, staying as close to that work as possible, only changing the input and output layers to match our input and output representations. \remitem{In each time step, we evaluate a dataset of tens of thousands of consumers. For this reason we use an approximate version of the Euclidean distance to speed up the clustering step. Instead of taking the real space for each dimension, we divide each dimension in a discrete number of bins. This effectively reduces the number of actual datapoints, since some consumers may be in the same bin in every dimension. This also allows a more efficient calculation of the Euclidean distance.}

\begin{figure*}
    \centering
    \includegraphics[width=\textwidth]{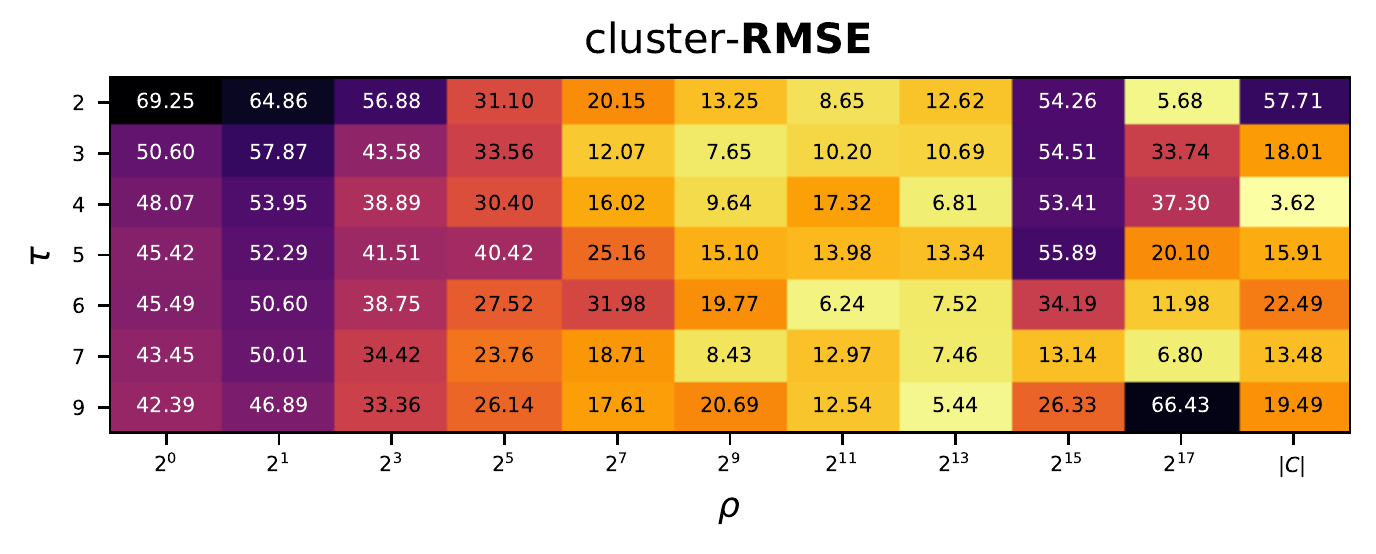}
    \caption{Results for the cluster-$\mathbf{RMSE}$. Lighter colours indicate a better (lower) value, the values are in monetary units.}
    \label{fig:res:clusterrmse}
\end{figure*}

\begin{figure*}
    \centering
    \includegraphics[width=\textwidth]{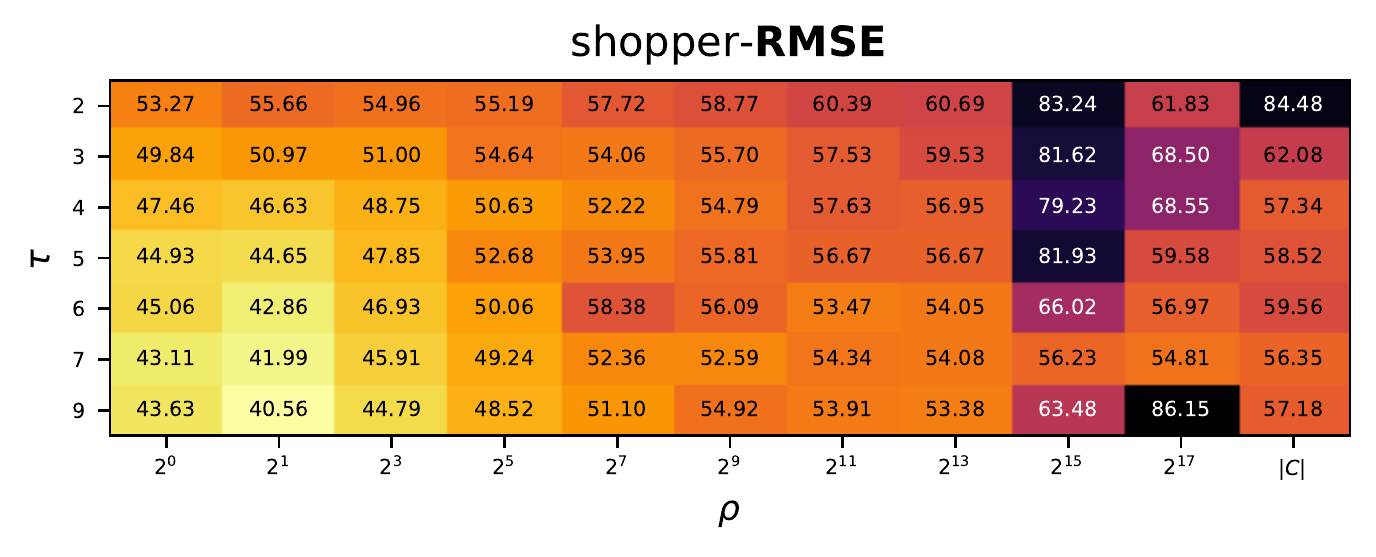}
    \caption{Results for the shopper-$\mathbf{RMSE}$. Lighter colours indicate a better (lower) value, the values are in monetary units.}
    \label{fig:res:shopperrmse}
\end{figure*}

\edititem{Each prediction estimates the spend of a proxy shopper, the average spend of a shopper in the respective cluster, which is then also used as prediction for each individual shopper in that cluster. Using the metrics defined in \cref{sec:framework:predict} we asses the quality of \fw for each combination of the $\tau$ and $\rho$ values. We compute the metrics at every time step, and average them over time. The results are shown in \crefrange{fig:res:clusterrmse}{fig:res:turnoverdropf1}.}{Each prediction estimates the average turnover per shopper in each cluster, which is subsequently assigned to the prediction for each shopper. We use these predictions to evaluate the framework as function of $\rho$ and $\tau$ at every time step $t$. \textit{Explanation of metrics moved to framework}Each of the four metrics is computed for all time steps, averaged over all time, and reported per combination of $\tau$ and $\rho$. Each combination of $\rho$ and $\tau$ results in one value for cluster-$\mathbf{RMSE}$, shopper-$\mathbf{RMSE}$, $\mathbf{F_1}$ and $\mathbf{APE}$, averaged over time, and presented in Figures \ref{fig:res:clusterrmse} - \ref{fig:res:turnoverdropf1} respectively.}

The results of each metric are presented as a heatmap, where better values (lower $\mathbf{RMSE}$ and $\mathbf{APE}$, higher $\mathbf{F_1}$) have a brighter colour, and worse values a darker one. Each column contains a single value of $\rho$ (increasing from left to right) and each row contains a single value of $\tau$ (increasing from top to bottom).

We answer the following questions:
\begin{enumerate}
\item How does $\tau$ influence the $\mathbf{RMSE}$ (prediction accuracy)?
\item How does $\rho$ influence the cluster-$\mathbf{RMSE}$ (prediction accuracy)?
\item How does $\rho$ influence the shopper-$\mathbf{RMSE}$ (accuracy and usefulness)?
\item How does $\rho$ influence the $\mathbf{APE}$ in $\hat{T}$ (usefulness)?
\item How does $\rho$ influence the $\mathbf{F_1}$ on the top decile (usefulness)?
\item How does $\rho$ influence the $\mathbf{F_1}$ \textit{over time} (usefulness)?
\end{enumerate}

In each of Sections \ref{sec:eeshopper:taurmse} through \ref{sec:eeshopper:rhof1time} we answer each of the above questions.

\subsubsection{\texorpdfstring{The effect of $\tau$ on $\mathbf{RMSE}$ (prediction accuracy)}{The effect of tau (prediction accuracy)}}
\label{sec:eeshopper:taurmse}
We first analyze the effects of $\tau$ on the cluster- and shopper-$\mathbf{RMSE}$ in Figures \ref{fig:res:clusterrmse} and \ref{fig:res:shopperrmse} respectively, as this will be relevant during the discussions on $\rho$. We distinguish between $\rho < 2^{15}$ and $\rho \geq 2^{15}$, where $\rho=\size{\entityset}$ is considered part of the latter.

$\mathbf{\rho <2^{15}}:$ For smaller values of $\rho$, all experiments show a clear decrease of $\mathbf{RMSE}$ with increasing $\tau$ for both $\mathbf{RMSE}$ results. This is expected, as a higher $\tau$ means that the data sequences for each cluster are longer and the LSTM can learn from a longer period of time, which benefits its performance \cite{Wen2016}.

$\mathbf{\rho \geq 2^{15}}:$ For higher $\rho$, the $\mathbf{RMSE}$ is not as consistently decreasing with increasing $\tau$. The reason for this is that for a higher value of $\rho$ we have fewer clusters of shoppers and hence fewer training points. This makes the model possibly less stable, as it has less data to improve its performance. As a result, some models may fail to perform as expected. This means that longer sequences (higher $\tau$) may result in worse predictions than shorter sequences (lower $\tau$).

\subsubsection{\texorpdfstring{The effect of $\rho$ on cluster-$\mathbf{RMSE}$ (prediction accuracy)}{The effect of rho on cluster-RMSE (prediction accuracy)}}
We next analyze the influence of $\rho$ (left to right) on the cluster-$\mathbf{RMSE}$, presented in \cref{fig:res:clusterrmse}. Up to $\rho=2^{15}$ we see a clear patterns: increasing $\rho$ decreases the clusters-$\mathbf{RMSE}$. This makes sense, as we increase the number of shoppers in each cluster. Because we are comparing the prediction of the average shopper with the average ground truth, we make a more accurate prediction for the shoppers \textit{as a group}. Put differently, some shoppers will have a lower ground truth than the predicted cluster average, others will have a higher ground truth than the cluster average. These two effects cancel each other to get a lower cluster-$\mathbf{RMSE}$, provided that the model can make a decent prediction in the first place. As argued in the analysis of $\tau$, this does not seem to apply for $\rho \geq 2^{15}$, where the above pattern is not guaranteed.

\subsubsection{\texorpdfstring{The effect of $\rho$ on the shopper-$\mathbf{RMSE}$}{The effect of rho on the shopper-RMSE}}
\label{sec:eeshopper:shopperrmse}
Contrary to the results for the cluster-$\mathbf{RMSE}$, \cref{fig:res:shopperrmse} shows that lower values of $\rho$ tend to produce better results for the shopper-$\mathbf{RMSE}$. Increasing $\rho$ from $2^0=1$ (no clustering) to $2^1=2$, we first see a small improvement when combining an average of two shoppers into a cluster, followed by a gradual increase of the shopper-$\mathbf{RMSE}$ as $\rho$ increases. This different result is caused by two factors. The first is that a lower value of $\rho$ means that fewer shoppers are together in the cluster, which means that the individual shoppers are likely closer to each other and as such to the average shopper. This means the average prediction can be a better indicator for the individual shoppers, also resulting in an average prediction that is closer to the individual ground truths. The opposite happens for larger $\rho$ where individual shoppers have more distance to the average shopper, and so are the individual ground truths to the predicted average. The second factor is that, as discussed previously, the cluster-$\mathbf{RMSE}$ benefits from the averaging effect, which allows over- and under-predictions to cancel each other. For the shopper-$\mathbf{RMSE}$, over- and under-predictions are both penalized without cancelling each other, increasing the $\mathbf{RMSE}$.

\begin{figure*}[t]
    \centering
    \includegraphics[width=\textwidth]{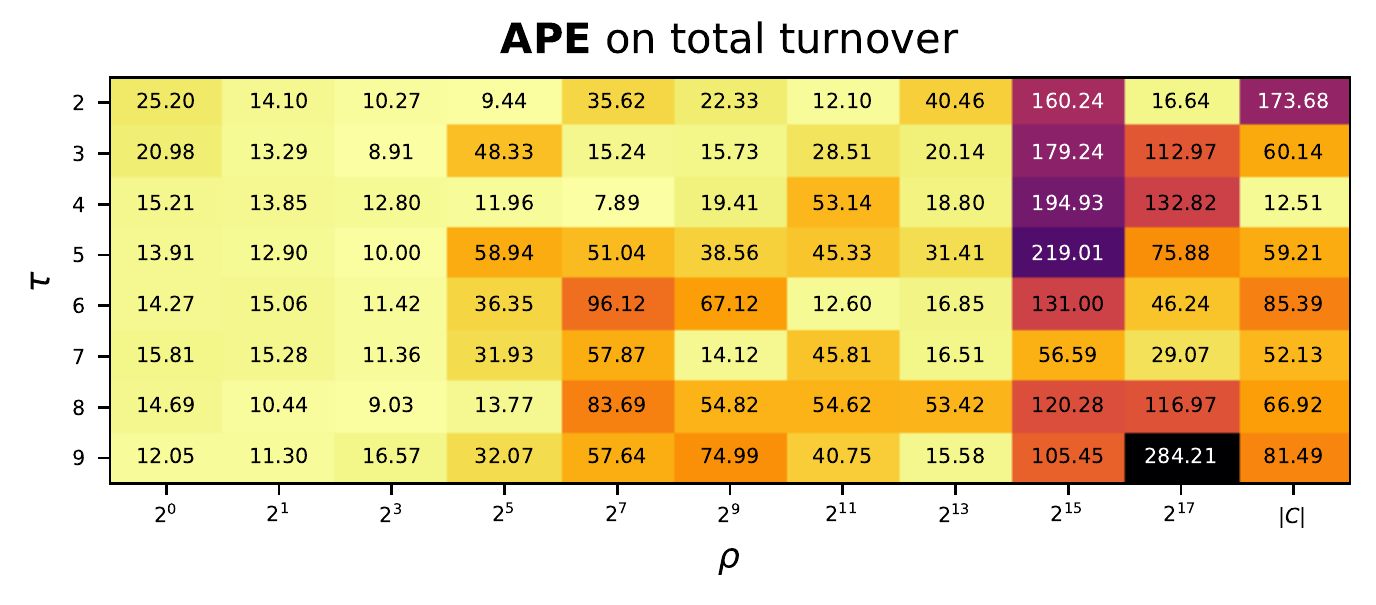}
    \caption{Results for the turnover $\mathbf{APE}$. Lighter colours indicate a better (lower) value, the values are error percentages with respect to the actual turnover.}
    \label{fig:res:turnoverape}
\end{figure*}

\subsubsection{\texorpdfstring{The effect of $\rho$ on $\mathbf{APE}$}{The effect of rho on APE}}
As depicted in Figure \ref{fig:res:turnoverape}, we see a similar relation between $\rho$ and the $\mathbf{APE}$ on the turnover. What is notable though, is that for medium high values of $\rho$ (between $2^5$ and $2^{13}$), some results are closer to the low values ($\leq 2^{5}$), indicated by the lighter colours. This is in contrast to the results of the shopper-$\mathbf{RMSE}$, where the results for these $\rho$ values are much worse (indicated by the darker colours for that range). Similar to the better results on the cluster-$\mathbf{RMSE}$, predicting the total turnover allows under- and over-predictions to cancel. This allows some of the mid-range $\rho$ to still reasonably predict the total turnover, despite the worse shopper-$\mathbf{RMSE}$.

\begin{figure*}
    \centering
    \includegraphics[width=\textwidth]{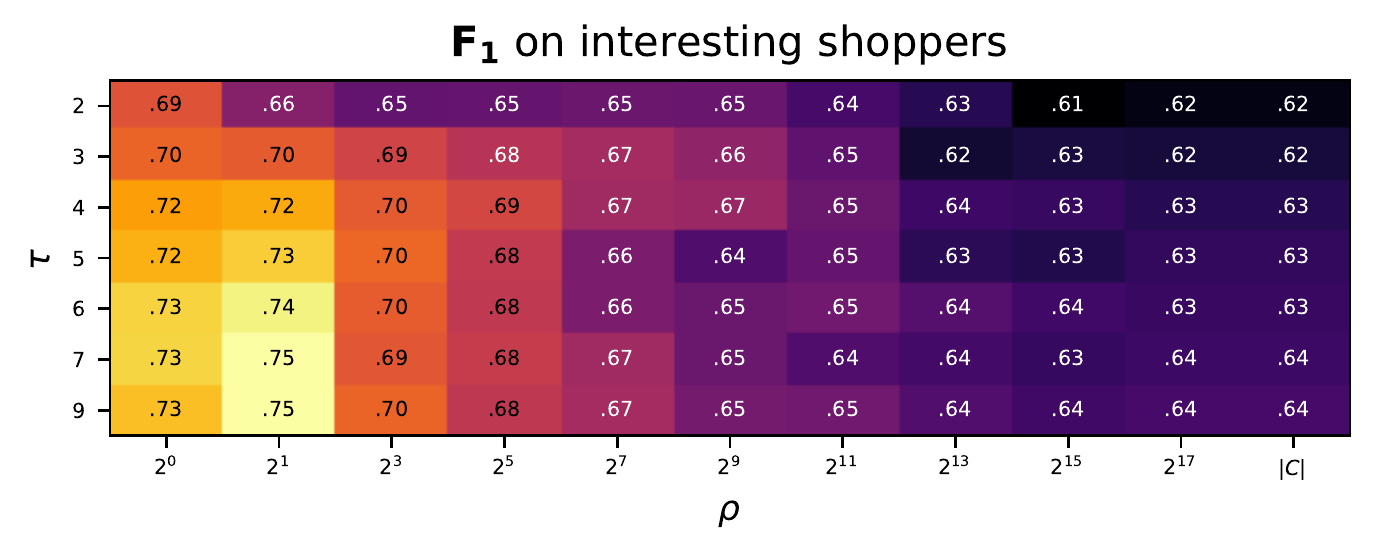}
    \caption{Results for the $\mathbf{F_1}$ on the shopper decile with the highest drop in turnover. Lighter colours indicate a better (higher) value.}
    \label{fig:res:turnoverdropf1}
\end{figure*}

\subsubsection{\texorpdfstring{The effect of $\rho$ on $\mathbf{F_1}$}{The effect of rho on F1}}
\label{sec:eeshoppers:rhof1}
We next discuss the effect of $\rho$ on the $\mathbf{F_1}$ metric when predicting the top decile of shoppers that lower their spend from one week to the next. Similar to the results for the shopper-$\mathbf{RMSE}$ in Section \ref{sec:eeshopper:shopperrmse}, the quality of the results decreases (lower $\mathbf{F_1}$) with increasing $\rho$. While this is partly to be expected given the increase in shopper-$\mathbf{RMSE}$ for higher $\rho$, this metric punishes for incorrect under/over predictions. A shopper with the ground truth of $10$ monetary units lower that the predicted value will contribute in the same way to the shopper-$\mathbf{RMSE}$ as a shopper with $10$ units under-prediction. This is not true for the metric presented in \cref{fig:res:turnoverdropf1}, it captures whether the error is, to a certain extend, consistent between shoppers as well\additem{, so over-predicting all shoppers by the same amount is less penalizing}. As discussed, being able to predict which shoppers will decrease their spend most from one week to the next is a measure of the usefulness. We see that being closer to the individual prediction (lower $\rho$) is better, though there is a small advantage of clustering with two shoppers per cluster.

\subsubsection{\texorpdfstring{The effect of $\rho$ on $\mathbf{F_1}$ over time}{The effect of rho on F1 over time}}
\label{sec:eeshopper:rhof1time}
\begin{figure*}
    \centering
    \includegraphics[width=\textwidth]{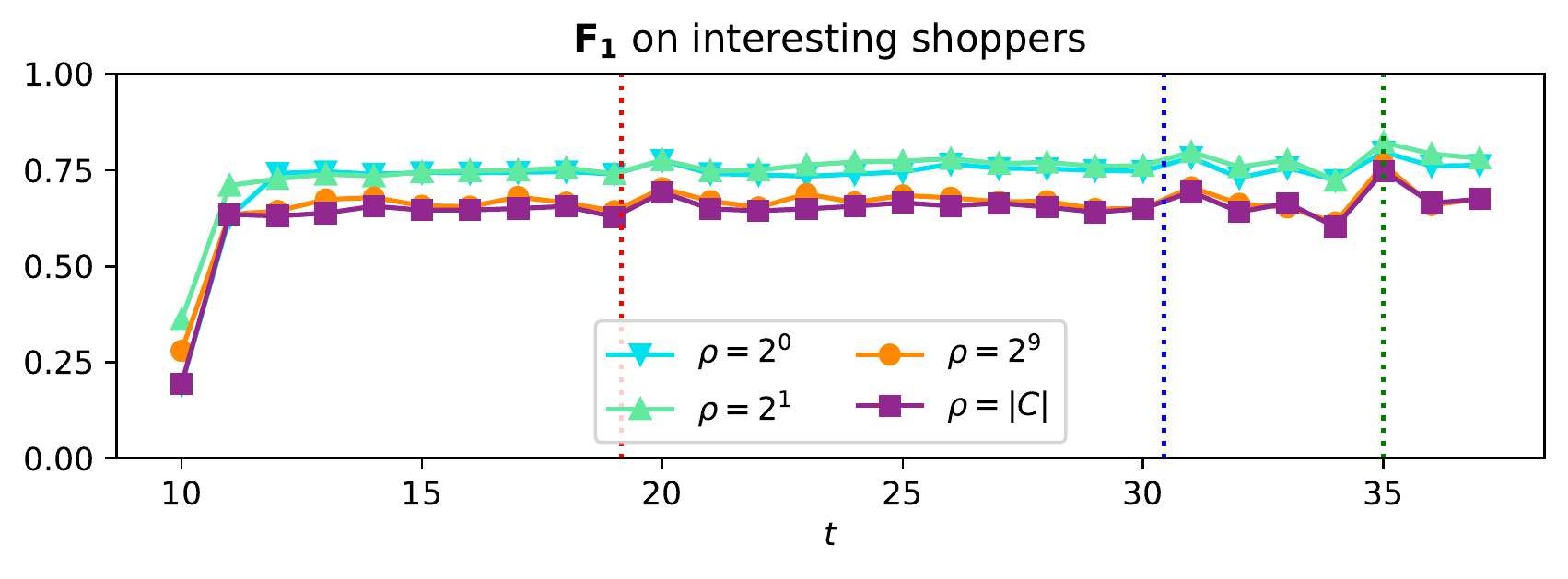}
    \caption{Progression of $\mathbf{F_1}$ over time for $\tau=9$. The vertical lines indicate significant external influences that can influence shopper behaviour.}
    \label{fig:res:time}
\end{figure*}

The results above have shown the effect of $\rho$ on prediction accuracy and usefulness as average over all time steps. We now discuss some of the effects that can be seen as \fw progresses over time, presented in Figure \ref{fig:res:time}. The numbers in the $\tau=9$ row of Figure \ref{fig:res:turnoverdropf1} are the averages of these plots. From the picture we see again that lower values of $\rho$ prevail, but that the progression of $\mathbf{F_1}$ over time is similar. \edititem{Note that for each value of $\rho$, $\mathbf{F_1}$ of the first time step is relatively low. This is because the first timestep has a cold-start, the prediction model needs data over more time to reach a steady level,}{ Each experiments needs one period to start up, but then remains steady over time,} with the higher values of $\rho$ settling about $0.1$ lower than the lower values of $\rho$. Next to this, we see the effect of external events at three points in time, indicated by vertical dotted lines. While we cannot disclose the details of the timeline, the third of these is very likely to temporarily change shopper behaviour. The other lines describe similar though less significant external effects. Because of these points of sudden changes in the shopper behaviour one expects a decrease in the predictive accuracy in the time around them. This results in a sudden drop in $\mathbf{F_1}$, especially for the first and third point in time, but these recover quickly. \additem{Of special interest is the ability of the model to handle concept drifts does not seem to be influenced by the value of $\rho$. In other words, any effect concept drift is shown to have on the prediction without \fw ($\rho=2^0$) is not worsened or improved by \fw. The progression of $\mathbf{F_1}$ over time is very similar over all values of $\rho$. Major deviations from the stable $\mathbf{F_1}$ occur at the same time. This is partly due to the incremental nature of the LSTM model. At the same time this is not trivial, as a higher $\rho$ means fewer datapoints to train on, and as such fewer datapoints to correct for the concept drifts. A similar argument can be made for the variety in seasons; the half a year of data does not show seasonal effects on the $\mathbf{F_1}$.}

\subsection{Paint Factory Experiments}\label{sec:results:paintfactory}
For the second use case, we use the BPIC 2019 dataset \cite{bpic2019} as depicted in Figure \ref{fig:processoverview}, for which the invoice features are summarized in \cref{tab:orderfeatures}. We first filter the dataset keeping only invoices with exactly one occurrence of `Vendor creates invoice' and `Record invoice receipt', in the correct order. We also remove cases that started before 2018 or ended after 2018. Furthermore, we only keep the invoice attributes described in \cref{tab:orderfeatures}. After this filtering, we have $171517$ invoices containing a total of $1025949$ events with  $\size{\activityspace}=40$ different labels. We discuss the experimental setup in \cref{sec:eepaint:setup} and present and discuss the results in \crefrange{sec:eepaint:results}{sec:eepaint:random}.

\subsubsection{Experimental Setup}\label{sec:eepaint:setup}
In this experiment we compare the error, expressed by $\mathbf{RMSE}$, \additem{between the predicted invoice duration $\hat{y}(c,t)$ and the actual invoice duration $y(c,t)$}, against the value of $\rho$. For the experiment we vary $\rho$, increasing from $1$ to $178$ in steps of $1$, the final value being just below the maximum value of $\size{\entityset_t^P}$, and $\rho=\size{\entityset}$. We therefore have a total of $179$ experiments. This higher resolution for $\rho$ (compared to Section \ref{sec:results:shoppers}) is possible because the dataset \edititem{has fewer events and each entity is only used once for prediction and training}{is much smaller}. Next to this, we repeat all experiments, but assign cases to \textit{random} clusters. This is done to compare the effect `averaging' in general with carefully selected clusters of similar entities. \additem{For the prediction model we use the Adaptive Hoeffding Option Tree, as implemented in \cite{skmultiflow}. This is different from the LSTM used for the shopper dataset, as we want to show that \fw works for different prediction models, as those are only a choice within \fw.}

\begin{figure*}
    \centering
    \includegraphics[width=\textwidth]{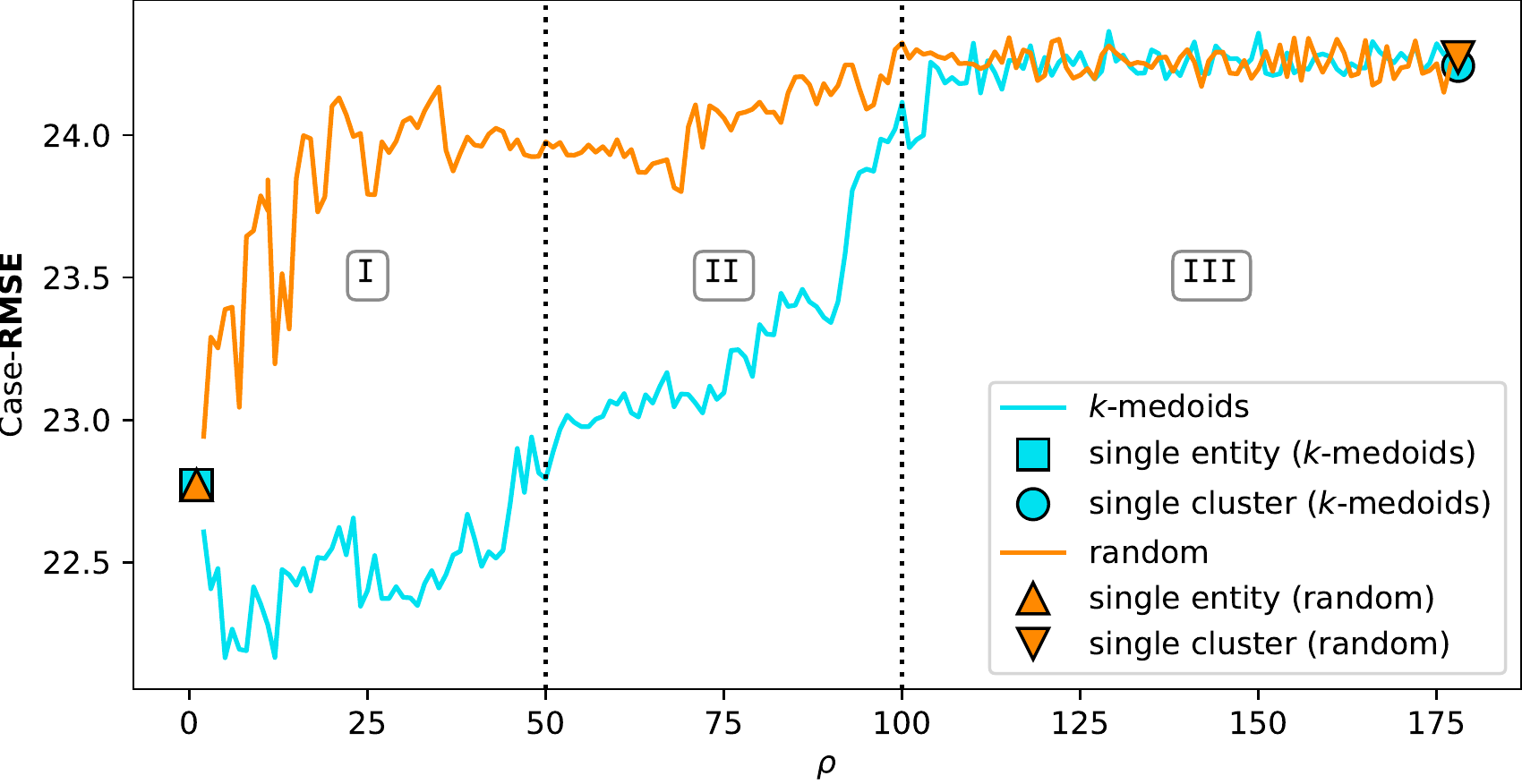}
    \caption{Results for the BPIC'19 event log. The blue line/symbols indicate the results for assigning entities to clusters using $k$-medoids and Gower's distance. The orange line/symbols indicate the results for randomly assigned groups.}
    \label{fig:res:events}
\end{figure*}

\subsubsection{Results}\label{sec:eepaint:results}
The results for the BPIC'19 log are presented in \cref{fig:res:events}. The plot shows two colours: the blue indicates the results for clustering with Gower's distance and $k$-medoids, and in orange the results for randomly assigning the entities to one of $k$ clusters. In both, $k$ is based on $\rho$ and $\size{\entityset_t^P}$ or $\size{\entityset_t^T}$ as per \cref{eq:rho}. The blue square and orange upward triangle indicate the special value $\rho=1$ (no clustering). The blue circle and orange downward triangle indicate the special value $\rho=\size{\entityset}$ (all entities in a single cluster).

\subsubsection{\texorpdfstring{Influence of $\rho$ on $\mathbf{RMSE}$}{Influence of rho on RMSE}}
Looking at the blue line in \cref{fig:res:events}, we distinguish three sections for $\rho$: low values from $2$ to about $50$ (indicated by \texttt{I}), mid values from $50$ to $100$ (\texttt{II}) and high values above $100$ (\texttt{III}). For low values we see an improvement of \fw over single invoice predictions (the blue square). This is in line with the results for the shopper dataset, where we find that a limited average cluster size has a positive influence on the prediction accuracy. For the mid range, we see a steady increase in the $\mathbf{RMSE}$, also confirming our findings from Section \ref{sec:results:shoppers}. In this range the invoices in the clusters deviate too much from the cluster average, resulting in worse invoice-$\mathbf{RMSE}$ values. At some point, the number of clusters rarely exceeds $1$ as per \cref{eq:rho}, leading to results close to $\rho = \size{\entityset}$, decreasing performance further as we reach a limited number of training points in each time step.

\subsubsection{The effect of clustering versus random clusters}\label{sec:eepaint:random}
Given \fw, one can wonder what the contribution is of creating clusters of \textit{similar} entities, and to what extend \textit{any} averaging plays a role in the results of \fw. To look into this, we compare two sets of experiments: applying the $k$-medoids clustering with Gower's distance against assigning entities to groups at random. The latter of these methods creates $k$ clusters, and for each entity selects one of the clusters at random to be assigned to. This creates the results presented in orange in \cref{fig:res:events}. For low values of $\rho$ (up to 100) we see large differences between the clusters created with $k$-medoids and the cluster created at random, while for high values (from 100) this difference vanishes. Similar to the discussion in \cref{sec:eeshopper:shopperrmse}, having fewer invoices in a cluster means that the invoices will be closer to the average invoice, but only if clusters are defined as such. For random clusters, the invoices may just as well be close to each other (and to the average) as far away from each other (and from the average), with most clusters expected to be somewhere in between. As we increase the number of invoices in a cluster (higher $\rho$) we also increase the distance from the invoices to the average invoice.

\subsection{\texorpdfstring{Considerations for $\rho$}{considerations for rho}}
\label{sec:results:conclusion}
In this section we look back at the results from the two rather different datasets and discuss conclusions that hold for both datasets. Based on the results there are three considerations for the optimal value of $\rho$: \textit{Model stability}, \textit{Cluster stability}, and \textit{Cluster detail}. For Model stability, we require enough clusters to be found to create sufficient training data for a stable model. This effect is explicitly visible for $\rho > 2^{13}$ in the shopper dataset. For the Cluster stability, we find that increasing $\rho$ improves the prediction accuracy on clusters, i.e. how well are we able to predict the behaviour of the entities in a cluster \textit{as a group}. This is visible in \cref{fig:res:clusterrmse}. Finally, for Cluster detail we do not want $\rho$ to be too large, as we make our predictions more tailored to the individual, an effect clearly visible in Figures \ref{fig:res:shopperrmse} through \ref{fig:res:events}. These are three considerations that should be balanced and taken into account when deciding the best value of $\rho$ for a given use case and dataset.

In the original version of this work, the parameter of \fw was the number of clusters $k$. In other words, the number of clusters that we find at each time step is predefined as a parameter and does not depend on $\size{\entityset_t^P}$ and $\size{\entityset_t^T}$ as per \cref{eq:rho}. While this works fine for datasets where the number of entities is roughly similar over time (such as for the supermarket dataset), it poses difficulties when such a condition does not hold (such as for the paint factory dataset). The introduction of $\rho$ solved this issue. \additem{As a result, \fw presented in this paper, next to being more generic, delivers a more stable performance for the paint factory dataset. The results on the supermarket dataset are similar (except that an increase in $\rho$ is a decrease in $k$ and vice-versa).}

An interesting question for future work is to see how these conclusions change for different datasets, especially with the number of entities in the datasets. The introduction of $\rho$ in this paper to replace $k$ makes reasoning about this easier. If we increase the size of the dataset, we also increase $\size{\entityset_t^P}$ and $\size{\entityset_t^T}$. This has two effects on the clustering. First, we increase the number of clusters we find at each time step (as per \cref{eq:rho}), and second, the entities are likely to be more similar. The first effect is expected to benefit Model stability, removing the negative effects that high $\rho$ experience. For the second effect, consider the supermarket dataset. If we add transaction data from another 160000 shoppers, for the same value of $\rho$, each of the 160000 shoppers in the original dataset will be as close or closer to the shoppers in the cluster they are assigned. As a result, the average shopper in each cluster will be a better representative. For the same value of $\rho$, this will improve the Cluster stability while maintaining the same level of Cluster detail, resulting in an overall performance increase.

\section{Conclusion and Future Work}\label{sec:conclusion}
In this paper we proposed \fw to make predictions about future events of entities by combining the training and prediction data of similar entities. This aims to overcome the problem that single entities may show very different behaviour, making accurate predictions unstable and inaccurate. Combining entities is done using clustering, where the number of clusters is determined by the desired average number of entities in a cluster $\rho$, a parameter of \fw. $\rho$ introduces a trade-off: a higher $\rho$ results in fewer training points and as such poorer trained model, but because we have fewer and as such larger clusters, the behaviour of entities as a group is better averaged, increasing prediction accuracy. Next to this, a lower $\rho$ also means smaller clusters, and as such the predictions are closer to the individual entities. We applied \fw to two use cases: shopper behaviour in supermarkets and invoice throughput prediction in the ordering process of a paint factory. In the supermarket use case we showed the trade-off. Making small groups of shoppers increases the quality of down-stream prediction tasks over using individual shoppers, but moving to larger groups decreases overall prediction quality. In the paint factory ordering process, we see a similar trend. Clustering invoices in small groups improves the prediction performance initially, but the prediction quality deteriorates for larger clusters.

For future research, several directions can be identified.
The most important one is to see how the size of the dataset affects the considerations for the optimal number of entities per cluster, as discussed in \cref{sec:results:conclusion}. \additem{This includes applying \fw to datasets from different retailers, longer periods of time or more shoppers, but also the application in different use cases.}
\additem{Apart from different datasets and use cases, the effect of \fw on other prediction tasks within the supermarket use case can be looked into as well. One such example is basket analysis. While this paper focuses on the spend of shoppers as outcome (and the derived interesting-shopper outcome of \cref{sec:eeshoppers:rhof1,sec:eeshopper:rhof1time}), other work looks into which products are likely to be bought by a shopper \cite{Doan2019a} or when a shopper next makes a visit \cite{Chen2018}. \fw is not limited to spend as outcome per se, but an interesting future research direction is adapting the basket content prediction to be used with \fw.}
Another direction to consider is using event-oriented clustering methods like the ones explored in the field of process mining. \edititem{The applications of \fw in this paper do not make use of the individual events for clustering, but on the encoding learned from entities extracted from events. Alternatives are existing methods from the process mining field such as \cite{Leontjeva2016,bose2012} that apply their clustering on the sequences of events.}{In the current framework we use (linear fits of) feature data to define clusters of similar entities. \fw operates on events that belong to entities, existing clustering methods from the process mining field, such as [14,16], are alternatives to this.}
In \fw, the found clusters are used as a \textit{means to an end}. Put differently, the found clusters are only used to improve the predictions, but are not analyzed beyond this.
\edititem{Next to this, the analysis of how clusters progress over time can be an interesting result as well, particularly in the supermarket use case. For example, do shoppers change between clusters from one week to another, and to what extend do external effects (concept drift, discussed in \cref{sec:eeshopper:rhof1time}) influence this?}{They can also be investigated as stand-alone results to find out how (groups of) shoppers behave over time.}
Finally, at each time step the clusters are recomputed based on the most recent data of the entities. An extension lies in incorporating past information on clusters, such that longer-term similarities in entity behaviour can also be considered. However, this is specific to use cases like the supermarket, where entities are used for training and prediction on multiple time steps, unlike the paint factory.

\section*{Conflict of interest}
The authors declare that they have no conflicts of interest related to this work.

\bibliographystyle{sn-mathphys}%
\bibliography{main}


\begin{thebibliography}{45}
\ifx \bisbn   \undefined \def \bisbn  #1{ISBN #1}\fi
\ifx \binits  \undefined \def \binits#1{#1}\fi
\ifx \bauthor  \undefined \def \bauthor#1{#1}\fi
\ifx \batitle  \undefined \def \batitle#1{#1}\fi
\ifx \bjtitle  \undefined \def \bjtitle#1{#1}\fi
\ifx \bvolume  \undefined \def \bvolume#1{\textbf{#1}}\fi
\ifx \byear  \undefined \def \byear#1{#1}\fi
\ifx \bissue  \undefined \def \bissue#1{#1}\fi
\ifx \bfpage  \undefined \def \bfpage#1{#1}\fi
\ifx \blpage  \undefined \def \blpage #1{#1}\fi
\ifx \burl  \undefined \def \burl#1{\textsf{#1}}\fi
\ifx \doiurl  \undefined \def \doiurl#1{\url{https://doi.org/#1}}\fi
\ifx \betal  \undefined \def \betal{\textit{et al.}}\fi
\ifx \binstitute  \undefined \def \binstitute#1{#1}\fi
\ifx \binstitutionaled  \undefined \def \binstitutionaled#1{#1}\fi
\ifx \bctitle  \undefined \def \bctitle#1{#1}\fi
\ifx \beditor  \undefined \def \beditor#1{#1}\fi
\ifx \bpublisher  \undefined \def \bpublisher#1{#1}\fi
\ifx \bbtitle  \undefined \def \bbtitle#1{#1}\fi
\ifx \bedition  \undefined \def \bedition#1{#1}\fi
\ifx \bseriesno  \undefined \def \bseriesno#1{#1}\fi
\ifx \blocation  \undefined \def \blocation#1{#1}\fi
\ifx \bsertitle  \undefined \def \bsertitle#1{#1}\fi
\ifx \bsnm \undefined \def \bsnm#1{#1}\fi
\ifx \bsuffix \undefined \def \bsuffix#1{#1}\fi
\ifx \bparticle \undefined \def \bparticle#1{#1}\fi
\ifx \barticle \undefined \def \barticle#1{#1}\fi
\bibcommenthead
\ifx \bconfdate \undefined \def \bconfdate #1{#1}\fi
\ifx \botherref \undefined \def \botherref #1{#1}\fi
\ifx \url \undefined \def \url#1{\textsf{#1}}\fi
\ifx \bchapter \undefined \def \bchapter#1{#1}\fi
\ifx \bbook \undefined \def \bbook#1{#1}\fi
\ifx \bcomment \undefined \def \bcomment#1{#1}\fi
\ifx \oauthor \undefined \def \oauthor#1{#1}\fi
\ifx \citeauthoryear \undefined \def \citeauthoryear#1{#1}\fi
\ifx \endbibitem  \undefined \def \endbibitem {}\fi
\ifx \bconflocation  \undefined \def \bconflocation#1{#1}\fi
\ifx \arxivurl  \undefined \def \arxivurl#1{\textsf{#1}}\fi
\csname PreBibitemsHook\endcsname

\bibitem{Chamberlain2017}
\begin{botherref}
\oauthor{\bsnm{Chamberlain}, \binits{B.P.}},
\oauthor{\bsnm{Liu}, \binits{B.}},
\oauthor{\bsnm{Pagliari}, \binits{R.}},
\oauthor{\bsnm{Deisenroth}, \binits{M.P.}}:
{Customer Lifetime Value Prediction Using Embeddings}.
KDD 2017 ADS
\textbf{17}
(2017).
\doiurl{10.1145/3097983.3098123}
\end{botherref}
\endbibitem

\bibitem{Ishigaki2018}
\begin{barticle}
\bauthor{\bsnm{Ishigaki}, \binits{T.}},
\bauthor{\bsnm{Terui}, \binits{N.}},
\bauthor{\bsnm{Sato}, \binits{T.}},
\bauthor{\bsnm{Allenby}, \binits{G.M.}}:
\batitle{{Personalized market response analysis for a wide variety of products
  from sparse transaction data}}.
\bjtitle{International Journal of Data Science and Analytics}
\bvolume{5}(\bissue{4}),
\bfpage{233}--\blpage{248}
(\byear{2018}).
\doiurl{10.1007/S41060-018-0099-9}
\end{barticle}
\endbibitem

\bibitem{Mishra2017}
\begin{barticle}
\bauthor{\bsnm{Mishra}, \binits{S.}},
\bauthor{\bsnm{Leroy}, \binits{V.}},
\bauthor{\bsnm{Amer-Yahia}, \binits{S.}}:
\batitle{{Colloquial region discovery for retail products: discovery and
  application}}.
\bjtitle{International Journal of Data Science and Analytics}
\bvolume{4}(\bissue{1}),
\bfpage{17}--\blpage{34}
(\byear{2017}).
\doiurl{10.1007/S41060-017-0048-Z}
\end{barticle}
\endbibitem

\bibitem{DeCnudde2020}
\begin{barticle}
\bauthor{\bsnm{{De Cnudde}}, \binits{S.}},
\bauthor{\bsnm{Martens}, \binits{D.}},
\bauthor{\bsnm{Evgeniou}, \binits{T.}},
\bauthor{\bsnm{Provost}, \binits{F.}}:
\batitle{{A benchmarking study of classification techniques for behavioral
  data}}.
\bjtitle{International Journal of Data Science and Analytics}
\bvolume{9}(\bissue{2}),
\bfpage{131}--\blpage{173}
(\byear{2020}).
\doiurl{10.1007/S41060-019-00185-1}
\end{barticle}
\endbibitem

\bibitem{Francescomarino2019}
\begin{barticle}
\bauthor{\bsnm{Francescomarino}, \binits{C.D.}},
\bauthor{\bsnm{Dumas}, \binits{M.}},
\bauthor{\bsnm{Maggi}, \binits{F.M.}},
\bauthor{\bsnm{Teinemaa}, \binits{I.}}:
\batitle{{Clustering-Based Predictive Process Monitoring}}.
\bjtitle{IEEE Transactions on Services Computing}
\bvolume{12}(\bissue{6}),
\bfpage{896}--\blpage{909}
(\byear{2019}).
\doiurl{10.1109/TSC.2016.2645153}
\end{barticle}
\endbibitem

\bibitem{difrancescomarino2018}
\begin{botherref}
\oauthor{\bsnm{Francescomarino}, \binits{C.D.}},
\oauthor{\bsnm{Ghidini}, \binits{C.}},
\oauthor{\bsnm{Maggi}, \binits{F.M.}},
\oauthor{\bsnm{Rizzi}, \binits{W.}},
\oauthor{\bsnm{Persia}, \binits{C.D.}},
\oauthor{\bsnm{Apr}, \binits{A.I.}}:
{Incremental Predictive Process Monitoring: How to Deal with the Variability of
  Real Environments}.
Technical report
(2018).
\url{https://arxiv.org/abs/1804.03967}
\end{botherref}
\endbibitem

\bibitem{Medeiros2007}
\begin{barticle}
\bauthor{\bparticle{de} \bsnm{Medeiros}, \binits{A.K.A.}},
\bauthor{\bsnm{Guzzo}, \binits{A.}},
\bauthor{\bsnm{Greco}, \binits{G.}},
\bauthor{\bparticle{van~der} \bsnm{Aalst}, \binits{W.M.P.}},
\bauthor{\bsnm{Weijters}, \binits{A.J.M.M.}},
\bauthor{\bsnm{{Van Dongen}}, \binits{B.F.}},
\bauthor{\bsnm{Sacc{\`{a}}}, \binits{D.}}:
\batitle{{Process Mining Based on Clustering: A Quest for Precision}}.
\bjtitle{Lecture Notes in Computer Science (including subseries Lecture Notes
  in Artificial Intelligence and Lecture Notes in Bioinformatics)}
\bvolume{4928 LNCS},
\bfpage{17}--\blpage{29}
(\byear{2007}).
\doiurl{10.1007/978-3-540-78238-4_4}
\end{barticle}
\endbibitem

\bibitem{Tax2017}
\begin{bchapter}
\bauthor{\bsnm{Tax}, \binits{N.}},
\bauthor{\bsnm{Verenich}, \binits{I.}},
\bauthor{\bsnm{{La Rosa}}, \binits{M.}},
\bauthor{\bsnm{Dumas}, \binits{M.}}:
\bctitle{{Predictive Business Process Monitoring with LSTM Neural Networks}}.
In: \beditor{\bsnm{Dubois}, \binits{E.}},
\beditor{\bsnm{Pohl}, \binits{K.}} (eds.)
\bbtitle{Advanced Information Systems Engineering},
pp. \bfpage{477}--\blpage{492}.
\bpublisher{Springer},
\blocation{Cham}
(\byear{2017}).
\doiurl{10.1007/978-3-319-59536-8_30}
\end{bchapter}
\endbibitem

\bibitem{Taymouri2021}
\begin{botherref}
\oauthor{\bsnm{Taymouri}, \binits{F.}},
\oauthor{\bsnm{{La Rosa}}, \binits{M.}},
\oauthor{\bsnm{Erfani}, \binits{S.M.}}:
{A Deep Adversarial Model for Suffix and Remaining Time Prediction of Event
  Sequences}.
Proceedings SDM 2021,
522--530
(2021).
\doiurl{10.1137/1.9781611976700.59}
\end{botherref}
\endbibitem

\bibitem{Leontjeva2016}
\begin{barticle}
\bauthor{\bsnm{Leontjeva}, \binits{A.}},
\bauthor{\bsnm{Conforti}, \binits{R.}},
\bauthor{\bsnm{Francescomarino}, \binits{C.D.}},
\bauthor{\bsnm{Dumas}, \binits{M.}},
\bauthor{\bsnm{Maggi}, \binits{F.M.}}:
\batitle{{Complex Symbolic Sequence Encodings for Predictive Monitoring of
  Business Processes}}.
\bjtitle{Lecture Notes in Computer Science (including subseries Lecture Notes
  in Artificial Intelligence and Lecture Notes in Bioinformatics)}
\bvolume{9253},
\bfpage{297}--\blpage{313}
(\byear{2016}).
\doiurl{10.1007/978-3-319-23063-4_21}
\end{barticle}
\endbibitem

\bibitem{Teinemaa2016}
\begin{barticle}
\bauthor{\bsnm{Teinemaa}, \binits{I.}},
\bauthor{\bsnm{Dumas}, \binits{M.}},
\bauthor{\bsnm{Maggi}, \binits{F.M.}},
\bauthor{\bsnm{Francescomarino}, \binits{C.D.}}:
\batitle{{Predictive Business Process Monitoring with Structured and
  Unstructured Data}}.
\bjtitle{Lecture Notes in Computer Science (including subseries Lecture Notes
  in Artificial Intelligence and Lecture Notes in Bioinformatics)}
\bvolume{9850 LNCS},
\bfpage{401}--\blpage{417}
(\byear{2016}).
\doiurl{10.1007/978-3-319-45348-4_23}
\end{barticle}
\endbibitem

\bibitem{Teinemaa2019}
\begin{barticle}
\bauthor{\bsnm{Teinemaa}, \binits{I.}},
\bauthor{\bsnm{Dumas}, \binits{M.}},
\bauthor{\bsnm{Rosa}, \binits{M.L.}},
\bauthor{\bsnm{Maggi}, \binits{F.M.}}:
\batitle{{Outcome-Oriented Predictive Process Monitoring: Review and
  Benchmark}}.
\bjtitle{ACM Trans. Knowl. Discov. Data}
\bvolume{13}(\bissue{2}),
\bfpage{1}--\blpage{57}
(\byear{2019}).
\doiurl{10.1145/3301300}
\end{barticle}
\endbibitem

\bibitem{Hassani2021}
\begin{barticle}
\bauthor{\bsnm{Hassani}, \binits{M.}},
\bauthor{\bsnm{Habets}, \binits{S.}}:
\batitle{{Predicting next touch point in a customer journey: A use case in
  telecommunication}}.
\bjtitle{Proceedings - European Council for Modelling and Simulation, ECMS}
\bvolume{35}(\bissue{1}),
\bfpage{48}--\blpage{54}
(\byear{2021}).
\doiurl{10.7148/2021-0048}
\end{barticle}
\endbibitem

\bibitem{Gunesen2021}
\begin{bchapter}
\bauthor{\bsnm{G{\"{u}}nesen}, \binits{S.N.}},
\bauthor{\bsnm{Şen}, \binits{N.}},
\bauthor{\bsnm{Yıldırım}, \binits{N.}},
\bauthor{\bsnm{Kaya}, \binits{T.}}:
\bctitle{{Customer Churn Prediction in FMCG Sector Using Machine Learning
  Applications}}.
In: \beditor{\bsnm{Mercier-Laurent}, \binits{E.}},
\beditor{\bsnm{Kayalica}, \binits{M.{\"{O}}.}},
\beditor{\bsnm{Owoc}, \binits{M.L.}} (eds.)
\bbtitle{Artificial Intelligence for Knowledge Management},
pp. \bfpage{82}--\blpage{103}.
\bpublisher{Springer},
\blocation{Cham}
(\byear{2021}).
\doiurl{10.1007/978-3-030-80847-1_6}
\end{bchapter}
\endbibitem

\bibitem{Tariq2021}
\begin{barticle}
\bauthor{\bsnm{Tariq}, \binits{M.U.}},
\bauthor{\bsnm{Babar}, \binits{M.}},
\bauthor{\bsnm{Poulin}, \binits{M.}},
\bauthor{\bsnm{Khattak}, \binits{A.S.}}:
\batitle{{Distributed model for customer churn prediction using convolutional
  neural network}}.
\bjtitle{Journal of Modelling in Management}
(\byear{2021}).
\doiurl{10.1108/JM2-01-2021-0032}
\end{barticle}
\endbibitem

\bibitem{Spenrath2020}
\begin{bchapter}
\bauthor{\bsnm{Spenrath}, \binits{Y.}},
\bauthor{\bsnm{Hassani}, \binits{M.}},
\bauthor{\bparticle{van} \bsnm{Dongen}, \binits{B.}},
\bauthor{\bsnm{Tariq}, \binits{H.}}:
\bctitle{{Why Did My Consumer Shop? Learning an Efficient Distance Metric for
  Retailer Transaction Data}}.
In: \bbtitle{Proceedings of ECML PKDD 2020 Lecture Notes in Computer Science
  Springer},
pp. \bfpage{323}--\blpage{338}
(\byear{2021}).
\doiurl{10.1007/978-3-030-67670-4_20}
\end{bchapter}
\endbibitem

\bibitem{Burattin2014}
\begin{bchapter}
\bauthor{\bsnm{Burattin}, \binits{A.}},
\bauthor{\bsnm{Sperduti}, \binits{A.}},
\bauthor{\bparticle{van~der} \bsnm{Aalst}, \binits{W.M.P.}}:
\bctitle{{Control-flow discovery from event streams}}.
In: \bbtitle{2014 IEEE Congress on Evolutionary Computation (CEC)},
pp. \bfpage{2420}--\blpage{2427}
(\byear{2014}).
\doiurl{10.1109/CEC.2014.6900341}
\end{bchapter}
\endbibitem

\bibitem{Spenrath2019}
\begin{botherref}
\oauthor{\bsnm{Spenrath}, \binits{Y.}},
\oauthor{\bsnm{Hassani}, \binits{M.}}:
{Ensemble-Based Prediction of Business Process Bottlenecks With Recurrent
  Concept Drifts}.
Workshop proceedings of the EDBT/ICDT 2019 Joint Conference
(2019)
\end{botherref}
\endbibitem

\bibitem{Bose2014}
\begin{barticle}
\bauthor{\bsnm{Bose}, \binits{R.P.J.C.}},
\bauthor{\bparticle{van~der} \bsnm{Aalst}, \binits{W.M.P.}},
\bauthor{\bsnm{{\v{Z}}liobaitė}, \binits{I.}},
\bauthor{\bsnm{Pechenizkiy}, \binits{M.}}:
\batitle{{Dealing With Concept Drifts in Process Mining}}.
\bjtitle{IEEE Transactions on Neural Networks and Learning Systems}
\bvolume{25}(\bissue{1}),
\bfpage{154}--\blpage{171}
(\byear{2014}).
\doiurl{10.1109/TNNLS.2013.2278313}
\end{barticle}
\endbibitem

\bibitem{Chen2018}
\begin{bchapter}
\bauthor{\bsnm{Chen}, \binits{T.}},
\bauthor{\bsnm{Keng}, \binits{B.}},
\bauthor{\bsnm{Moreno}, \binits{J.}}:
\bctitle{{Multivariate Arrival Times with Recurrent Neural Networks for
  Personalized Demand Forecasting}}.
In: \bbtitle{2018 IEEE International Conference on Data Mining Workshops
  (ICDMW)},
pp. \bfpage{810}--\blpage{819}
(\byear{2018}).
\doiurl{10.1109/ICDMW.2018.00121}
\end{bchapter}
\endbibitem

\bibitem{Doan2019a}
\begin{barticle}
\bauthor{\bsnm{Doan}, \binits{T.}},
\bauthor{\bsnm{Veira}, \binits{N.}},
\bauthor{\bsnm{Keng}, \binits{B.}}:
\batitle{{Generating Realistic Sequences of Customer-level Transactions for
  Retail Datasets}}.
\bjtitle{IEEE ICDM Workshop on Data Mining for Services 2018}
(\byear{2019}).
\doiurl{10.48550/arXiv.1901.05577}
\end{barticle}
\endbibitem

\bibitem{Guidotti2021}
\begin{bchapter}
\bauthor{\bsnm{Guidotti}, \binits{R.}},
\bauthor{\bsnm{Nanni}, \binits{M.}},
\bauthor{\bsnm{Giannotti}, \binits{F.}},
\bauthor{\bsnm{Pedreschi}, \binits{D.}},
\bauthor{\bsnm{Bertoli}, \binits{S.}},
\bauthor{\bsnm{Speciale}, \binits{B.}},
\bauthor{\bsnm{Rapoport}, \binits{H.}}:
\bctitle{{Measuring Immigrants Adoption of Natives Shopping Consumption with
  Machine Learning}}.
In: \bbtitle{Lecture Notes in Computer Science (including Subseries Lecture
  Notes in Artificial Intelligence and Lecture Notes in Bioinformatics)}
vol. \bseriesno{12461 LNAI},
pp. \bfpage{369}--\blpage{385}
(\byear{2021}).
\doiurl{10.1007/978-3-030-67670-4_23}
\end{bchapter}
\endbibitem

\bibitem{VanderAalst2016}
\begin{bbook}
\bauthor{\bparticle{van~der} \bsnm{Aalst}, \binits{W.M.P.}}:
\bbtitle{Process Mining},
\bedition{2nd} edn.
\bpublisher{Springer},
\blocation{Berlin, Heidelberg}
(\byear{2016}).
\doiurl{10.1007/978-3-662-49851-4}
\end{bbook}
\endbibitem

\bibitem{Ariannezhad2021}
\begin{bchapter}
\bauthor{\bsnm{Ariannezhad}, \binits{M.}},
\bauthor{\bsnm{Jullien}, \binits{S.}},
\bauthor{\bsnm{Nauts}, \binits{P.}},
\bauthor{\bsnm{Fang}, \binits{M.}},
\bauthor{\bsnm{Schelter}, \binits{S.}},
\bauthor{\bparticle{de} \bsnm{Rijke}, \binits{M.}}:
\bctitle{{Understanding Multi-Channel Customer Behavior in Retail}}.
In: \bbtitle{Proceedings of the 30th ACM International Conference on
  Information \& Knowledge Management},
pp. \bfpage{2867}--\blpage{2871}.
\bpublisher{Association for Computing Machinery},
\blocation{New York, NY, USA}
(\byear{2021}).
\burl{https://doi.org/10.1145/3459637.3482208}
\end{bchapter}
\endbibitem

\bibitem{Bai2019}
\begin{bchapter}
\bauthor{\bsnm{Bai}, \binits{G.-J.}},
\bauthor{\bsnm{Lien}, \binits{C.-Y.}},
\bauthor{\bsnm{Chen}, \binits{H.-H.}}:
\bctitle{{Co-Learning Multiple Browsing Tendencies of a User by Matrix
  Factorization-Based Multitask Learning}}.
In: \bbtitle{IEEE/WIC/ACM International Conference on Web Intelligence}.
\bsertitle{WI '19},
pp. \bfpage{253}--\blpage{257}.
\bpublisher{Association for Computing Machinery},
\blocation{New York, NY, USA}
(\byear{2019}).
\doiurl{10.1145/3350546.3352526}
\end{bchapter}
\endbibitem

\bibitem{Unnikrishnan2020}
\begin{barticle}
\bauthor{\bsnm{Unnikrishnan}, \binits{V.}},
\bauthor{\bsnm{Beyer}, \binits{C.}},
\bauthor{\bsnm{Matuszyk}, \binits{P.}},
\bauthor{\bsnm{Niemann}, \binits{U.}},
\bauthor{\bsnm{Pryss}, \binits{R.}},
\bauthor{\bsnm{Schlee}, \binits{W.}},
\bauthor{\bsnm{Ntoutsi}, \binits{E.}},
\bauthor{\bsnm{Spiliopoulou}, \binits{M.}}:
\batitle{{Entity-level stream classification: exploiting entity similarity to
  label the future observations referring to an entity}}.
\bjtitle{International Journal of Data Science and Analytics}
\bvolume{9}(\bissue{1}),
\bfpage{1}--\blpage{15}
(\byear{2020}).
\doiurl{10.1007/S41060-019-00177-1}
\end{barticle}
\endbibitem

\bibitem{Ferriera2009}
\begin{bchapter}
\bauthor{\bsnm{Ferreira}, \binits{D.R.}},
\bauthor{\bsnm{Gillblad}, \binits{D.}}:
\bctitle{{Discovering Process Models from Unlabelled Event Logs}}.
In: \beditor{\bsnm{Dayal}, \binits{U.}},
\beditor{\bsnm{Eder}, \binits{J.}},
\beditor{\bsnm{Koehler}, \binits{J.}},
\beditor{\bsnm{Reijers}, \binits{H.A.}} (eds.)
\bbtitle{Business Process Management},
pp. \bfpage{143}--\blpage{158}.
\bpublisher{Springer},
\blocation{Berlin, Heidelberg}
(\byear{2009}).
\doiurl{10.1007/978-3-642-03848-8_11}
\end{bchapter}
\endbibitem

\bibitem{Bayomie2016}
\begin{barticle}
\bauthor{\bsnm{Bayomie}, \binits{D.}},
\bauthor{\bsnm{Helal}, \binits{I.M.A.}},
\bauthor{\bsnm{Awad}, \binits{A.}},
\bauthor{\bsnm{Ezat}, \binits{E.}},
\bauthor{\bsnm{ElBastawissi}, \binits{A.}}:
\batitle{{Deducing Case IDs for Unlabeled Event Logs}}.
\bjtitle{Lecture Notes in Business Information Processing}
\bvolume{256},
\bfpage{242}--\blpage{254}
(\byear{2016}).
\doiurl{10.1007/978-3-319-42887-1_20}
\end{barticle}
\endbibitem

\bibitem{Song2009}
\begin{bchapter}
\bauthor{\bsnm{Song}, \binits{M.}},
\bauthor{\bsnm{G{\"{u}}nther}, \binits{C.W.}},
\bauthor{\bparticle{van~der} \bsnm{Aalst}, \binits{W.M.P.}}:
\bctitle{{Trace Clustering in Process Mining}}.
In: \bbtitle{Lecture Notes in Business Information Processing},
vol. \bseriesno{17},
pp. \bfpage{109}--\blpage{120}
(\byear{2009}).
\doiurl{10.1007/978-3-642-00328-8_11}
\end{bchapter}
\endbibitem

\bibitem{bose2012}
\begin{botherref}
\oauthor{\bsnm{{Jagadeesh Chandra Bose}}, \binits{R.P.}}:
{Process mining in the large}.
PhD thesis,
Mathematics and Computer Science
(2012).
\doiurl{10.6100/IR730954}
\end{botherref}
\endbibitem

\bibitem{Lloyd1982}
\begin{barticle}
\bauthor{\bsnm{Lloyd}, \binits{S.P.}}:
\batitle{{Least Squares Quantization in PCM}}.
\bjtitle{IEEE Transactions on Information Theory}
\bvolume{28}(\bissue{2}),
\bfpage{129}--\blpage{137}
(\byear{1982}).
\doiurl{10.1109/TIT.1982.1056489}
\end{barticle}
\endbibitem

\bibitem{Spenrath2022}
\begin{bchapter}
\bauthor{\bsnm{Spenrath}, \binits{Y.}},
\bauthor{\bsnm{Hassani}, \binits{M.}},
\bauthor{\bsnm{{Van Dongen}}, \binits{B.F.}}:
\bctitle{{BitBooster: Effective Approximation of Distance Metrics via Binary
  Operations}}.
In: \bbtitle{Proceedings of COMPSAC 2022}
(\byear{2022})
\end{bchapter}
\endbibitem

\bibitem{VandenBerg2021}
\begin{bchapter}
\bauthor{\bparticle{Van~den} \bsnm{Berg}, \binits{S.}},
\bauthor{\bsnm{Hassani}, \binits{M.}}:
\bctitle{{On Inferring a Meaningful Similarity Metric for Customer Behaviour}}.
In: \bbtitle{Proceedings of ECML PKDD 2021 Lecture Notes in Computer Science
  Springer},
pp. \bfpage{234}--\blpage{250}
(\byear{2021}).
\doiurl{10.1007/978-3-030-86517-7_15}
\end{bchapter}
\endbibitem

\bibitem{Ester1996}
\begin{bchapter}
\bauthor{\bsnm{Ester}, \binits{M.}},
\bauthor{\bsnm{Kriegel}, \binits{H.-P.}},
\bauthor{\bsnm{Sander}, \binits{J.}},
\bauthor{\bsnm{Xu}, \binits{X.}}:
\bctitle{{A Density-Based Algorithm for Discovering Clusters in Large Spatial
  Databases with Noise}}.
In: \bbtitle{Proceedings of the Second International Conference on Knowledge
  Discovery and Data Mining}.
\bsertitle{KDD'96},
pp. \bfpage{226}--\blpage{231}
(\byear{1996}).
\burl{https://dl.acm.org/doi/10.5555/3001460.3001507}
\end{bchapter}
\endbibitem

\bibitem{Lin2017}
\begin{barticle}
\bauthor{\bsnm{Lin}, \binits{W.-C.}},
\bauthor{\bsnm{Tsai}, \binits{C.-F.}},
\bauthor{\bsnm{Hu}, \binits{Y.-H.}},
\bauthor{\bsnm{Jhang}, \binits{J.-S.}}:
\batitle{{Clustering-based undersampling in class-imbalanced data}}.
\bjtitle{Information Sciences}
\bvolume{409-410},
\bfpage{17}--\blpage{26}
(\byear{2017}).
\doiurl{10.1016/j.ins.2017.05.008}
\end{barticle}
\endbibitem

\bibitem{Garcia2006}
\begin{bchapter}
\bauthor{\bsnm{Baena-Garc{\'{i}}a}, \binits{M.}},
\bauthor{\bsnm{Campo-{\'{A}}vila}, \binits{J.}},
\bauthor{\bsnm{Fidalgo-Merino}, \binits{R.}},
\bauthor{\bsnm{Bifet}, \binits{A.}},
\bauthor{\bsnm{Gavald{\`{a}}}, \binits{R.}},
\bauthor{\bsnm{Morales-Bueno}, \binits{R.}}:
\bctitle{{Early Drift Detection Method}},
pp. \bfpage{77}--\blpage{86}
(\byear{2006}).
\burl{https://www.cs.upc.edu/
\end{bchapter}
\endbibitem

\bibitem{Bifet2009}
\begin{bchapter}
\bauthor{\bsnm{Bifet}, \binits{A.}},
\bauthor{\bsnm{Holmes}, \binits{G.}},
\bauthor{\bsnm{Pfahringer}, \binits{B.}},
\bauthor{\bsnm{Kirkby}, \binits{R.}},
\bauthor{\bsnm{Gavald{\`{a}}}, \binits{R.}}:
\bctitle{{New Ensemble Methods for Evolving Data Streams}}.
In: \bbtitle{Proceedings of the 15th ACM SIGKDD International Conference on
  Knowledge Discovery and Data Mining}.
\bsertitle{KDD '09},
pp. \bfpage{139}--\blpage{148}.
\bpublisher{ACM},
\blocation{New York, NY, USA}
(\byear{2009}).
\doiurl{10.1145/1557019.1557041}
\end{bchapter}
\endbibitem

\bibitem{skmultiflow}
\begin{barticle}
\bauthor{\bsnm{Montiel}, \binits{J.}},
\bauthor{\bsnm{Read}, \binits{J.}},
\bauthor{\bsnm{Bifet}, \binits{A.}},
\bauthor{\bsnm{Abdessalem}, \binits{T.}}:
\batitle{{Scikit-Multiflow: A Multi-output Streaming Framework}}.
\bjtitle{Journal of Machine Learning Research}
\bvolume{19}(\bissue{72}),
\bfpage{1}--\blpage{5}
(\byear{2018}).
\doiurl{10.5555/3291125.3309634}
\end{barticle}
\endbibitem

\bibitem{SOFTWARE:MOA}
\begin{bchapter}
\bauthor{\bsnm{Bifet}, \binits{A.}},
\bauthor{\bsnm{Holmes}, \binits{G.}},
\bauthor{\bsnm{Kirkby}, \binits{R.}},
\bauthor{\bsnm{Pfahringer}, \binits{B.}}:
\bctitle{{MOA: massive online analysis}},
vol. \bseriesno{11}
(\byear{2010}).
\burl{https://dl.acm.org/doi/10.5555/1756006.1859903}
\end{bchapter}
\endbibitem

\bibitem{Spenrath2020MT}
\begin{bchapter}
\bauthor{\bsnm{Spenrath}, \binits{Y.}},
\bauthor{\bsnm{Hassani}, \binits{M.}}:
\bctitle{{Predicting Business Process Bottlenecks In Online Events Streams
  Under Concept Drifts}}.
In: \beditor{\bsnm{Steglich}, \binits{M.}},
\beditor{\bsnm{Muller}, \binits{C.}},
\beditor{\bsnm{Neumann}, \binits{G.}},
\beditor{\bsnm{Walther}, \binits{M.}} (eds.)
\bbtitle{Proceedings of European Council for Modelling and Simulation (ECMS)
  2020}.
\bsertitle{Proceedings European Council for Modelling and Simulation},
pp. \bfpage{190}--\blpage{196}
(\byear{2020}).
\doiurl{10.7148/2020-0190}
\end{bchapter}
\endbibitem

\bibitem{Li2019}
\begin{botherref}
\oauthor{\bsnm{Li}, \binits{Z.}},
\oauthor{\bsnm{Xiong}, \binits{Y.}},
\oauthor{\bsnm{Huang}, \binits{W.}}:
{Drift-detection Based Incremental Ensemble for Reacting to Different Kinds of
  Concept Drift}
(2019).
\doiurl{10.1109/BIGCOM.2019.00025}
\end{botherref}
\endbibitem

\bibitem{bpic2019winners}
\begin{bchapter}
\bauthor{\bsnm{{Kiarash Diba}}, \binits{S.R.}},
\bauthor{\bsnm{Pufahl}, \binits{L.}}:
\bctitle{{BPI Challenge 2019: Performance and Compliance Analysis of
  Procurement Processes Using Process Mining}}.
In: \bbtitle{BPI Challenge}
(\byear{2019}).
\burl{https://icpmconference.org/2019/wp-content/uploads/sites/6/2019/07/BPI-Challenge-Submission-6.pdf}
\end{bchapter}
\endbibitem

\bibitem{bpic2019}
\begin{botherref}
\oauthor{\bsnm{Van~Dongen}, \binits{B.}}:
{BPI Challenge 2019}.
4TU.ResearchData
(2019).
\doiurl{10.4121/uuid:d06aff4b-79f0-45e6-8ec8-e19730c248f1}
\end{botherref}
\endbibitem

\bibitem{Gower1971}
\begin{barticle}
\bauthor{\bsnm{Gower}, \binits{J.C.}}:
\batitle{{A General Coefficient of Similarity and Some of Its Properties}}.
\bjtitle{Biometrics}
\bvolume{27}(\bissue{4}),
\bfpage{857}--\blpage{871}
(\byear{1971}).
\doiurl{10.2307/2528823}
\end{barticle}
\endbibitem

\bibitem{Wen2016}
\begin{bchapter}
\bauthor{\bsnm{Wen}, \binits{Y.}},
\bauthor{\bsnm{Zhang}, \binits{W.}},
\bauthor{\bsnm{Luo}, \binits{R.}},
\bauthor{\bsnm{Wang}, \binits{J.}}:
\bctitle{{Learning text representation using recurrent convolutional neural
  network with highway layers}}.
In: \bbtitle{Neu-IR Workshop Preceedings, SIGIR'16}
(\byear{2016}).
\doiurl{10.48550/arXiv.1606.06905}
\end{bchapter}
\endbibitem

\end{thebibliography}


\end{document}